\documentclass[sn-aps, iicol]{sn-jnl}

\usepackage{graphicx}%
\usepackage{multirow}%
\usepackage{amsmath,amssymb,amsfonts}%
\usepackage{amsthm}%
\usepackage{mathrsfs}%
\usepackage[title]{appendix}%
\usepackage{xcolor}%
\usepackage{textcomp}%
\usepackage{manyfoot}%
\usepackage{booktabs}%
\usepackage{algorithm}%
\usepackage{algorithmicx}%
\usepackage{algpseudocode}%
\usepackage{listings}%
\usepackage{makecell}

\usepackage[citestyle=numeric,sorting=nyt]{biblatex}

\usepackage{siunitx}
\usepackage{booktabs}
\usepackage{comment}
\usepackage{caption}
\usepackage{subcaption}
\usepackage{tikz}

\usepackage[edges]{forest}
\usetikzlibrary{arrows.meta}
\usetikzlibrary{shadows}

\usepackage{fontawesome}
\usepackage{hyperref}
\usepackage{lipsum}
\usepackage{pifont}
\usepackage{tabularx}
\usepackage[raggedrightboxes]{ragged2e}

\newcommand{\cmark}{\ding{51}}%
\newcommand{\xmark}{\ding{55}}%

\def\etal{~et~al.~}
\def\eg{e.g.,~}
\def\ie{i.e.,~}

\def\cf{cf.~}

\def\sota{state-of-the-art~}

\def\wrt{w.r.t~}

\theoremstyle{thmstyleone}%
\theoremstyle{thmstyletwo}%

\theoremstyle{thmstylethree}%

\begin{document}

\title[Article Title]{Semantically-aware Neural Radiance Fields for Visual Scene Understanding: A Comprehensive Review}



\author[1,*]{\fnm{Thang-Anh-Quan} \sur{Nguyen}}

\author[2,*]{\fnm{Amine} \sur{Bourki}}

\author[3]{\fnm{Mátyás} \sur{Macudzinski}}

\author[4]{\fnm{Anthony} \sur{Brunel}}

\author[5]{\fnm{Mohammed} \sur{Bennamoun}}

\affil[1]{\orgname{Noah’s Ark, Huawei Paris Research Center}, \orgaddress{\country{France}}}

\affil[2]{\orgname{Inception Lab}, \country{Paris, France}}


\affil[3]{\orgdiv{Centrale Lille}, \orgname{University of Lille}, \orgaddress{\country{France}}}


\affil[4]{\orgaddress{\country{Nice, France}}}

\affil[5]{\orgname{The University of Western Australia}, \orgaddress{\city{Perth}, \country{Australia}}}
\affil[]{\orgname{{*: Denotes co-primary authorship}}}



\abstract{
This review thoroughly examines the role of semantically-aware Neural Radiance Fields (NeRFs) in visual scene understanding, covering an analysis of over 250 scholarly papers. It explores how NeRFs adeptly infer 3D representations for both stationary and dynamic objects in a scene. This capability is pivotal for generating high-quality new viewpoints, completing missing scene details (inpainting), conducting comprehensive scene segmentation (panoptic segmentation), predicting 3D bounding boxes, editing 3D scenes, and extracting object-centric 3D models. A significant aspect of this study is the application of semantic labels as viewpoint-invariant functions, which effectively map spatial coordinates to a spectrum of semantic labels, thus facilitating the recognition of distinct objects within the scene. Overall, this survey highlights the progression and diverse applications of semantically-aware neural radiance fields in the context of visual scene interpretation.
}


\maketitle

\begin{table*}[htbp]
  \centering
  \resizebox*{0.96\linewidth}{!}{
  \centering
  \begin{tabularx}{1.3\textwidth}{p{1.7cm}|p{1.8cm}|p{1.5cm}|p{0.9cm}|X|X}
  
    \toprule
  
    \textbf{Survey} & \textbf{Venue} & {\textbf{Sem. Tasks}} & {\textbf{Sem. Focus}} & \textbf{Strengths} & \textbf{Limitations}\\
        \midrule
    Xia \textit{et al.}\cite{xia2023survey} & CS 2023 & ..E... & \textcolor{red}{\xmark} & {Extensively covers 3D-aware image synthesis, implicit scene representations, differentiable neural rendering in NeRF-based methods.} & {Broad scope in image synthesis with \textbf{little attention to semantics}. Includes generation and editability. Study \textbf{limited to December 2022}, before most of competitive SRF methods were published.}\\
        \midrule
    Zhu \textit{et al.}\cite{zhu2023deep} & APSIPA 2023 & {G.E...} & \textcolor{red}{\xmark} & {Provides synthetic overview of the field on a wide variety of topics with emphasis of current limitations of generic approaches.} & {Broad scope with \textbf{very superficial attention} to semantic aspects. \textbf{Study limited to December 2022}, before most of competitive SRF methods were published.}\\ 
        \midrule
    Tewari \textit{et al.}\cite{tewari2022advances} & CGF 2022 & {G.E.H.} & \textcolor{red}{\xmark} & {Comprehensively explains neural rendering and scene representation techniques, challenges and improvement strategies. Discusses editable and NeRFs and compositionality of its time. Extensive discussion on challenges and perspectives.} & {Broad scope with \textbf{very superficial attention to semantic aspects} (handful of papers), focuses on graphics and rendering related geometric aspects. Study \textbf{limited to 2021} papers before most of competitive SRF methods were published.}\\ 
            \midrule
    Xie \textit{et al.}\cite{xie2022neural} & CGF 2022  & G.E.H. & \textcolor{red}{\xmark} & {In depth review with a strong tutorial on theoretical aspects of differentiable rendering, NeRFs, and scene representations.} & {Broad scope focusing on appearance, textures and relighting applications with \textbf{very superficial attention} to semantic aspects. Study \textbf{limited to very early 2022}, before most competitive SRF methods were published.} \\
            \midrule
            \midrule
    Mittal \textit{et al.}\cite{mittal2023neural}* & arXiv 2023 & \textbf{GSEOHL} & \textcolor{red}{\xmark} & {Emphasizes on the basics. Lists out short abstracts of 500 papers and pre-prints. Organized by loss functions, applications, publication year. Up-to-date and gets incremental updates for each new NeRF publication.} & {It is a 400+ page unpublished tech report with comprehensive tutorials, \textbf{not a journal publication}. Mainly a \textbf{list of abstract summaries}, \textbf{lacks analysis, discussions, and global insights}. It is a \textbf{constantly updated work in progress}.}\\ 
            \midrule
    Li \textit{et al.}\cite{li2023generative} & arXiv 2023 & ..E..L & \textcolor{red}{\xmark} & {In-depth review of text-guided and text-controlled strategies, with applications on the generation of avatars, textures, scenes, and shapes.} & {Narrow scope \textbf{restricted to TextTo3D generative methods} and partially covers editability. Study limited to May 2023. \textbf{Only addresses a small fraction of our scope in time and applications}.}\\
            \midrule
    Rabby \textit{et al.}\cite{rabby2023beyondpixels} & arXiv 2023 & G.E.H. & \textcolor{red}{\xmark} & {Includes discussions on compositionality, scene editing, and public datasets. Provides a centralized benchmark.} & {Broad scope with very \textbf{superficial attention to semantics}. Study \textbf{limited to 2022 papers, before most of competitive SRF methods were published}.}\\ 
            \midrule
    Slapak \textit{et al.}\cite{vslapak2023neural}  & arXiv 2023 & G...H. & \textcolor{red}{\xmark} & {Good general overview on geometric approaches relevant to industrial and robotics fields. Discusses effiency and effectiveness improvements of traditional NeRFs.} & {Relatively short survey. Narrow scope \textbf{restricted to industrial and robotic applications}. Only partially and \textbf{superficially covers semantic aspects G and H}. Study limited to early 2023, missing many references from CVPR 2023 onwards.}\\
            \midrule
    Gao \textit{et al.}\cite{gao2022nerf} & arXiv 2022 &  GSE... & \textcolor{red}{\xmark} &  {Recent 2023 scope. Good generalist overview. Covers public datasets and evaluation.} & {Broad scope with \textbf{very superficial overview of a few SRFs (handful of papers)}. Focus is \textbf{diluted among many geometric and efficiency oriented non-semantic} discussions, leaving modest text real estate for semantics.}\\ 
    \midrule
    \midrule
    \textbf{Ours} & - & \textbf{GSEOHL} & \textcolor{green}{\cmark} & {\textbf{Comprehensive analysis on semantics, up to submission date in January 2024}.} & {Purposefully focuses on semantics.} \\ 
    \bottomrule
  \end{tabularx}
  }
    \caption{\textbf{Comparative overview of previously existing NeRF surveys \wrt semantics (SRFs).} Semantic Tasks  include: \underline{G:} 3D Geometry Enhancement, \underline{S:} Segmentation, \underline{E:} Editable NeRFs, \underline{O:} Object Detection and 6D Pose, \underline{H:} Holistic Decomposition, \underline{L:} NeRFs and Language, \underline{.:} denotes missing task. \underline{Semantic Focus} refers to whether the primary focus of the study is on semantics. \underline{*:} Interesting reference, but not a journal paper.}
  \label{tab:surveys_overview}
\end{table*}


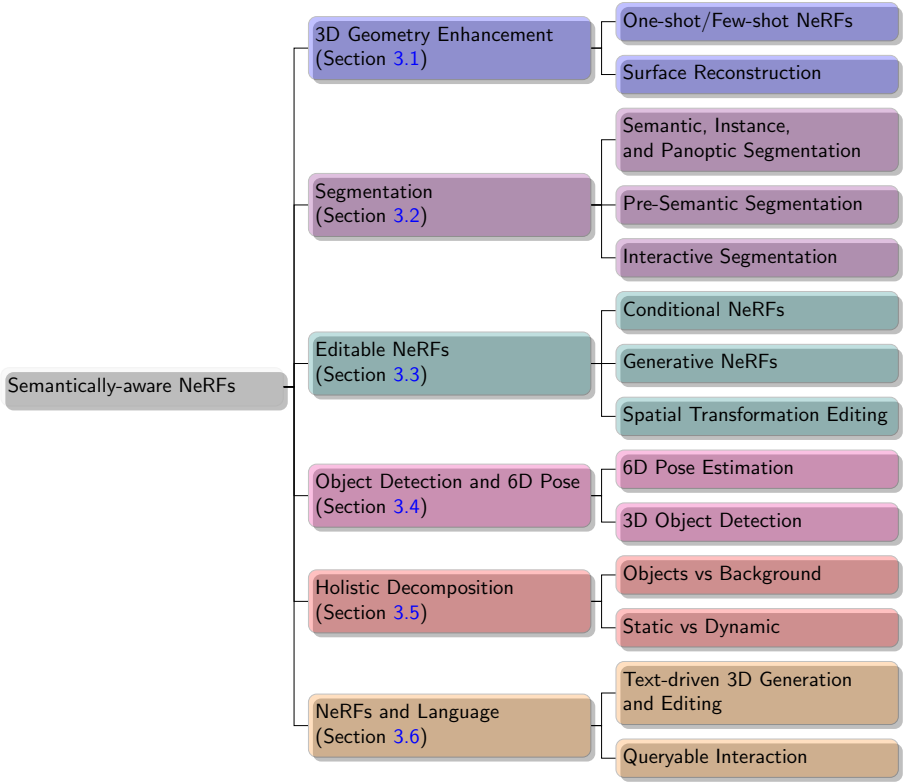
\begin{figure*}[htbp]
    \centering
    \resizebox{0.75\textwidth}{!}
    {
    \begin{forest}
          basic/.style = {draw, thin, drop shadow, font=\sffamily},
      upper style/.style = {basic, rounded corners=6pt, fill=black!6, text width=13.0em},
      lower style/.style = {basic, rounded corners=0pt, fill=black!10, text width=9em},
      where level<=2{%
        upper style,
        edge path'={
          (!u.parent anchor) -- +(0,-5pt) -| (.child anchor)
        },
      }{%
        lower style,
      },
      where level<=1{%
        parent anchor=children,
        child anchor=parent,
        if={isodd(n_children())}{%
          calign=child edge,
          calign primary child/.process={
            O+nw+n{n children}{(#1+1)/2}
          },
        }{%
          calign=edge midpoint,
        },
      }{
        folder,
        grow'=0,
      },
        for tree={
            grow'=0,
            rounded corners,
            align=left,draw,
            fill=white, 
            forked edges,
        },
        [Semantically-aware NeRFs, fill=gray, opacity=0.05
            [3D Geometry Enhancement\\(Section~\ref{sec:improve_3d}), fill=blue, opacity=0.25
                [One-shot/Few-shot NeRFs, fill=blue, opacity=0.25]
                [Surface Reconstruction, fill=blue, opacity=0.25]]
            [Segmentation\\(Section~\ref{sec:segmentation}), fill=violet, opacity=0.25
                [Semantic$\text{,}$ Instance$\text{,}$\\and Panoptic Segmentation, fill=violet, opacity=0.25]
                [Pre-Semantic Segmentation, fill=violet, opacity=0.25]
                [Interactive Segmentation, fill=violet, opacity=0.25]]
            [Editable NeRFs\\(Section~\ref{sec:edit_nerf}), fill=teal, opacity=0.25
                [Conditional NeRFs, fill=teal, opacity=0.25]
                [Generative NeRFs, fill=teal, opacity=0.25]
                [Spatial Transformation Editing, fill=teal, opacity=0.25]]
            [Object Detection and 6D Pose\\(Section~\ref{sec:obj_det}), fill=magenta, opacity=0.25
                [6D Pose Estimation, fill=magenta, opacity=0.25]
                [3D Object Detection, fill=magenta, opacity=0.25]]
            [Holistic Decomposition\\(Section~\ref{sec:decomposition}), fill=red, opacity=0.25
                [Objects vs Background, fill=red, opacity=0.25]
                [Static vs Dynamic, fill=red, opacity=0.25]]
            [NeRFs and Language\\(Section~\ref{sec:language}), fill=orange, opacity=0.25
                [Text-driven 3D Generation\\and Editing, fill=orange, opacity=0.25]
                [Queryable Interaction, fill=orange, opacity=0.25]]
        ]
    \end{forest}
    }
    \caption{{Taxonomy of our study on Semantically-aware Neural Radiance Fields (SRFs).}}
    \label{fig:taxonomy}
\end{figure*}

\section{Introduction}


Neural Radiance Fields (NeRFs) have marked a significant development since their inception~\cite{mildenhall2020nerf}, offering unprecedented capabilities in synthesizing photorealistic unseen views from a set of 2D images through a new 3D scene representation. The core strength of NeRFs lies in their ability to intricately model the complex interactions of light within a scene, thereby generating 3D representations that are both detailed and realistic. Traditional NeRFs, however, primarily focus on geometric and photometric accuracy, and often overlook the underlying semantics of the observed scenes.

The advent of semantically-aware NeRFs (SRFs) marks a significant advancement in this domain. These models not only capture the physical characteristics of a scene, but also incorporate an understanding of semantic and contextual information. This leap in technology facilitates a range of sophisticated applications, such as scene editing, improved object recognition, and more interactive and realistic virtual environments.


Recent developments in implicit neural rendering have also been pivotal. These methods demonstrate the possibility of learning accurate view synthesis for complex scenes by predicting their volumetric density and color, using only a set of RGB images as supervision. Despite these advances, most of the existing methods are limited to static scenes. They tend to encode all scene objects into a single neural network, thus falling short of representing dynamic scenes or breaking down scenes into individual objects. This limitation is a significant hurdle on the path towards creating more dynamic and responsive 3D environments.


Visual Scene Understanding, which is often categorized into the three R's of Computer Vision~\cite{malik2016three}, namely Reconstruction, Recognition, and Re-organization (\ie bottom-up segmentation), has received a massive attention in the field, both as individual problems as well as joint multitask approaches aiming to take advantage of their inherent mutually-informative nature, leading to improved performance and efficiency~\cite{atanov2022task,vandenhende2021multi,standley2020tasks,fifty2021efficiently,zamir2018taskonomy}. Similarly, the conventional methods which sequentiallly solve for NeRF then perception not only introduce additional computational costs and inefficiencies for perception tasks, but also fail to fully leverage the mutually-beneficial potential of volumetric renderings \wrt perception during the training phase. This gap represents a missed opportunity to maximize the synergies between 3D scene reconstruction and semantic perception.

Our comprehensive survey explores deep into these aspects, exploring the most recent advancements in semantically-aware NeRFs. We examine how the integration of semantic information can substantially enhance the capabilities of NeRFs, especially in complex and dynamic environments. Our discussion covers various methodologies for integrating semantic data into radiance fields, the challenges inherent in these processes, and the vast potential applications of these enriched models across diverse domains.

\textbf{Positioning and Impact.} The ultimate goal of this paper is to provide a thorough understanding of the current state and potential of semantically-aware NeRFs. We aim to identify gaps in existing methodologies, highlight the challenges yet to be overcome, and offer a vision for future research directions. To our knowledge, \textit{this survey is the first in the field to specifically concentrate on semantic coupling in Neural Radiance Fields}. This is significant considering the growing interest in this area of study.
%

\subsection{Prior Related Surveys}
This section focuses on previously conducted surveys that have explored Neural Radiance Fields (NeRFs), with a particular emphasis on semantic scene understanding from different perspectives, including 2D, 2.5D, and multi-view imaging techniques. These surveys have laid the groundwork in the field and provide insight into the development, capabilities, and limitations of NeRFs in processing and interpreting complex visual data.

Our survey expands upon this existing knowledge base by considering a wide range of venues and studies within a specific timeframe, offering a contemporary snapshot of the advancements in the field of semantically aware NeRFs. We not only review the findings and methodologies of these prior surveys but also highlight how our survey stands out in its approach and focus.

In particular, our survey explores how recent advancements in NeRF technology have been tailored to enhance semantic scene understanding. This includes exploring how these advanced NeRF models interpret and interact with complex visual scenes, pushing the boundaries of what is possible in terms of visual perception and scene interpretation. We also discuss the methodological approaches these studies have adopted, providing an analytical framework that contrasts our approach with those of previous surveys.

More specifically, in the survey published in November 2023, Xia\etal\cite{xia2023survey} extensively cover the field of 3D-aware image synthesis, including detailed discussions on implicit scene representations such as occupancy fields, signed distance fields, and radiance fields, with a particular emphasis on NeRFs. It also examines differentiable neural rendering, underscoring its crucial role in fine-tuning neural networks for 3D rendering and highlighting the importance of volume rendering in NeRF-based methods. However, compared to our survey, certain limitations become apparent.

The broader scope of~\cite{xia2023survey} contrasts with the more focused approach of our survey, which explores deeply the integration of semantic understanding into NeRFs. Our narrower focus allows for a more comprehensive exploration of how semantic integration can enhance or extend NeRFs, especially in complex and dynamic environments, which is an aspect that may not be covered as thoroughly in~\cite{xia2023survey}.

Furthermore, our survey potentially offers richer insights into the practical applications and challenges associated with implementing semantically enhanced NeRFs. These practical considerations are underrepresented in~\cite{xia2023survey}. In terms of future research directions, our survey provides more targeted guidance specific to semantic understanding in NeRFs, whereas~\cite{xia2023survey} may present a wider range of future trends and research areas across the broader field of 3D-aware image synthesis.

In summary, while both surveys make significant contributions to the field of Computer Vision and 3D Image Synthesis, our survey stands out for its specialized and in-depth focus on the semantic aspects of Neural Radiance Fields, offering nuanced perspectives and insights that are particularly relevant to the advancement of semantic integration in this area.

In a tech report also from November 2023, Gao\etal\cite{gao2022nerf} offer a broad overview of Neural Radiance Fields, discussing NeRF models, training requirements, various datasets used in research, and quality assessment metrics such as PSNR, SSIM, and LPIPS. However, compared to our survey, theirs shows limitations in its focus and depth. Our survey specifically concentrates on the integration of semantic understanding into NeRFs, providing detailed insights into semantic enhancement in complex and dynamic environments, practical applications, and specific future research directions. In contrast,~\cite{gao2022nerf} covers a wider range of topics in NeRF but lacks the specialized focus on semantic aspects, making our survey more comprehensive and targeted toward advancing semantic integration in Neural Radiance Fields.

In contrast to published surveys which typically cover NeRFs, as outlined in Table~\ref{tab:surveys_overview}, our goal is to provide readers with a thorough understanding of semantically-aware NeRFs research. By comparing our approach with previous surveys, we aim to emphasize the distinct contributions and insights our study provides, especially regarding the incorporation and interpretation of semantic information within the Neural Radiance Fields framework.

\subsection{Scope and Methodology}
The papers referenced in this survey are predominantly published in the top venues for Computer Vision, Computer Graphics, Machine Learning, and Robotics. They cover the period from 2020 (first NeRF paper~\cite{mildenhall2020nerf}) to the submission of the present paper in January 2024.

Our study primarily focuses on a taxonomy 6 main categories of semantically-aware NeRFs that define the underlying notion of semantics considered in this work, as summarized in Figure~\ref{fig:taxonomy}. First, we consider 3D geometry approaches that primarily use semantic information to improve performance in geometry-oriented tasks such as novel view synthesis and surface reconstruction. Specifically, in the very challenging setups of one-to-few shot scenarios, \ie with a very limited number of input views, NeRF-based methods can cope with the underconstrained nature of such challenging settings by leveraging higher-level information.
In addition to `reconstruction' applications, our study also includes segmentation which considers both the `recognition' and `re-organization' R's of Visual Scene Understanding (resp. semantic and pre-semantic segmentations)~\cite{malik2016three}. Editable NeRFs allow to manipulate scenes through various priors and strategies. We also discuss works that enrich a radiance field formulation with 3D Object Detection or 6D Pose considerations. Holistic decomposition which aims at encoding the exhaustive structure of an input scene in a top-down manner. Lastly, we study language-rich NeRFs that enable new multi-modal applications for human interaction or effective scene manipulation.

Semantics in 3D visual computing applications has been thoroughly explored with often vastly differing definitions and considerations. In the context of this study, our intended definition of `semantics' can be dissected into three main categories. {\bf Initially}, we consider semantics as an explicit higher level construct to designate object- and / or instance-level labels~\cite{siddiqui2023panoptic}, 3D object bounding boxes~\cite{hu2023nerf}, 3D object 6D poses~\cite{rozumnyi2023tracking}, or scene-wide decomposition~\cite{ost2021neural,liang2023semantic}. {\bf Secondly}, as for Editable NeRFs and certain 3D Geometry Enhancement methods, we consider the use of semantics through the lens of language representation learning (\eg \cite{wang2022clip}) which aim to describe scene objects with compact, controllable codes~\cite{jang2021codenerf}. This is typically used in order to efficiently improve multi-view consistency, cope with missing views~\cite{jain2021putting}, or to enable object- or scene level manipulation~\cite{muller2022autorf}. {\bf Finally}, we consider SRF strategies that explicitly interconnect vision and language, in order to generate novel 3D contents through text prompts~\cite{kerr2023lerf}, or enable higher-level scene interactions and user-guided manipulation~\cite{shen2023F3RM}.

Our work is also intended to help Computer Vision researchers unfamiliar with the topic step into semantically-aware NeRFs. Therefore, we cover the key concepts of the original NeRF architecture~\cite{mildenhall2020nerf} and a classic extension for jointly considering semantic segmentation~\cite{zhi2021place}. Additionally, we present a comprehensive overview of the relevant public datasets and evaluation tools. This additionally includes a centralized view of leading methods on these public benchmarks, by grouping results that are initially scattered across dozens of referenced papers, and presenting original discussions and insights.

\begin{figure*}[!htbp]
    \centering
    \includegraphics[width=\linewidth]{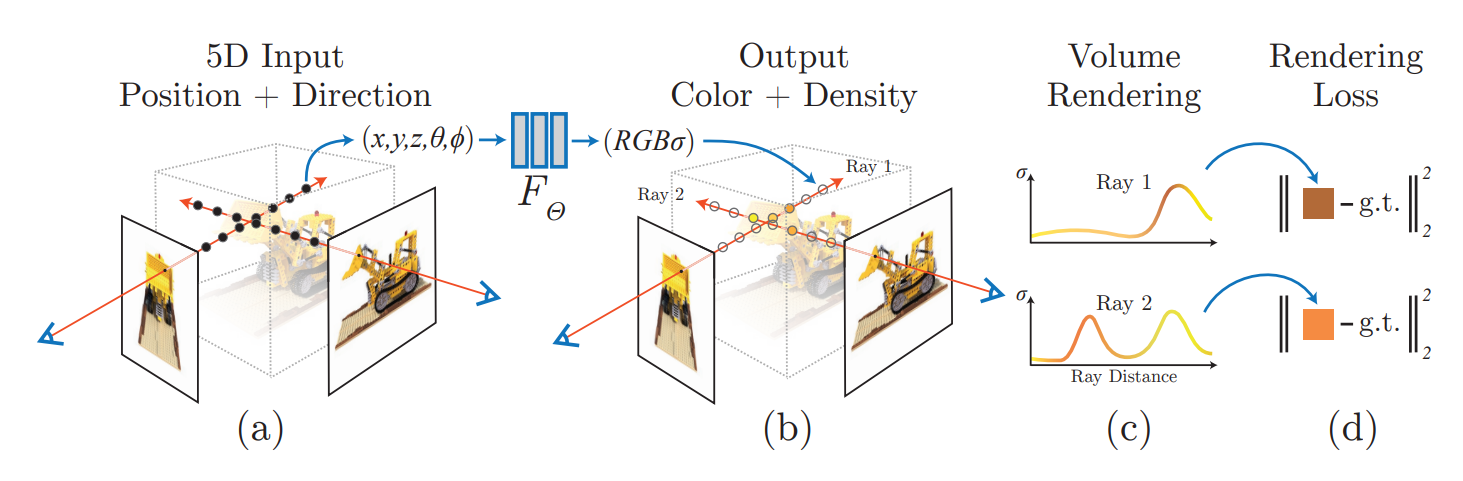}
    \caption{\textbf{Overview of NeRF~\cite{mildenhall2020nerf} scene representation and differentiable rendering.} (a) Images are synthesized by sampling 5D coordinates (location and viewing direction) along camera rays, (b) an MLP produces a color and volume density from those sampled points, and (c) volume rendering allow to reconstruct the final image using those values, all of which is end-to-end differentiable (d).}
    \label{fig:nerf_pipeline}
\end{figure*}

\subsection{Organization of the Paper}

The remainder of this paper is organized as follows. Section~\ref{sec:fundamentals} discusses the key principles of the standard NeRF formulation and its extension to a basic semantic task, \ie semantic segmentation. Section~\ref{sec:review} provides an extensive literature review of SRFs, while Section~\ref{sec:evaluation} reviews the main public datasets, metrics, and evaluation tools commonly used in this field. Section~\ref{sec:challenges} explores current challenges and perspectives, highlighting potential improvements in understanding semantic scenes and exploring real-world applications. Finally, 
Section~\ref{sec:conclusion} concludes the paper, giving higher-level perspectives for the field. 
We will also maintain an in-depth project repository on GitHub at \href{https://github.com/abourki/SoTA-Semantically-aware-NeRFs}{\textcolor{magenta}{github.com/abourki/SoTA-Semantically-aware-NeRFs}}, including a comprehensive list of references, datasets, and performance evaluations, with regular updates to provide the latest \sota developments.
%

\section{Fundamentals of Neural Radiance Fields}
\label{sec:fundamentals}
This section presents the core principles and terminology involved in the initial NeRF paper, as well as one of its simpler extensions to incorporate semantic reasoning capabilities.
To do so, we cover how the 3D scene is represented along the formal definition of NeRFs, how they are employed to generate novel views. For more general or geometry-oriented details, we refer the interested reader to other existing surveys that emphasize more on such aspects than our study, which focuses on semantic considerations, \eg\cite{xia2023survey,tewari2022advances,xie2022neural,gao2022nerf}.

\subsection{Scene Representation and Problem Statement}
Neural Radiance Fields (NeRFs), introduced by Mildenhall\etal~\cite{mildenhall2020nerf}, have revolutionized the field of novel view synthesis. A NeRF model encapsulates a 3D scene through a radiance field, which is essentially a 5D function that describes the light intensity traversing every direction within the scene. This is achieved by specifying both the color (as RGB values) and the volume density at each point in space. The core of a NeRF model lies in its ability to approximate this radiance function using Multi-Layer Perceptrons (MLPs). In the standard NeRF framework~\cite{mildenhall2020nerf}, a single MLP, denoted $F_\Theta$, is used for this purpose, as follows:
%

\begin{equation}
    (\mathbf{c}, \sigma) = F_\Theta(\mathbf{x}, \mathbf{d})
\end{equation}
where $\mathbf{x} = (x, y, z)$ is a given 3D point with $x, y, z$ coordinates, $\mathbf{d} = (\theta, \phi)$ represents the viewing direction in Euler angles, $\mathbf{c} = (r, g, b)$ the color, and $\sigma$ the corresponding volume density.  

\subsubsection{3D Scene Representation}
Radiance fields are typically encoded using either of two different approaches to representing 3D scenes: implicit and explicit representations, making them implicit or explicit radiance fields respectively. When using an implicit scene representation, \eg Signed Distance Functions (SDFs) or Deep Neural Networks (DNNs) in the case of NeRFs, the underlying geometry of the scene is not explicitly defined nor stored. It has to be retrieved using a post-processing or querying step, making then much more memory-efficient to the expense of additional computation.

Explicit Radiance Fields on the other hand rely on a data structure that explicitly defines the scene geometry as, \eg point clouds~\cite{huang2023boosting}, voxel grids~\cite{sun2022direct}, or permutohedral lattices~\cite{rosu2023permutosdf} that allow to store radiance information with faster access rates but often with scene resolution constraints linked to their superior memory complexity.

\subsubsection{Volumetric Rendering}
Volume rendering~\cite{kajiya1984ray} is a technique used to compute the color $C(r)$ of any camera ray $\mathbf{r}(t) = \mathbf{o} + t\mathbf{d}$ where $\mathbf{o}$ represents the camera position and $\mathbf{d}$ is the viewing direction, given the volume density and color functions of the scene being rendered. The color $C(r)$ is given by: 
%
\begin{equation}
    C(\mathbf{r}) = \sum_{i=1}^{N}T_i(1-\exp(-\sigma_i\delta_i))\mathbf{c}_i
\end{equation}
where $T_i=\exp{\left(-\sum_{j=1}^{i-1}\sigma_j\delta_j\right)}$ and $\delta_i = t_{i+1} - t_i$ is the distance between adjacent samples. This function is trivially differentiable and reduces to traditional alpha compositing with alpha values $\alpha_i = 1 - \exp(-\sigma_i\delta_i)$.

\subsubsection{Training for Novel View Synthesis}
During training, for each pixel, a square error photometric loss $\mathcal{L}_{color}$ is used to optimize the MLP parameters, as follows:

\begin{equation}
    \mathcal{L}_{color} = \frac{1}{|\mathcal{R}|}\sum_{r\in \mathcal{R}}\left\|\hat{C}(\mathbf{r}) - C(\mathbf{r})\right\|_2^2
\end{equation}
where $\mathcal{R}$ is the set of rays in each batch, and $C(\mathbf{r})$, $\hat{C}(\mathbf{r})$ are the ground truth, volume predicted RGB colors for ray $r$ respectively. The training procedure is typically scene-specific and requires dense images along their 3D poses and intrinsic parameters, and scene bounds which can be estimated using Structure-from-Motion (SfM) end-to-end frameworks, \eg COLMAP~\cite{schonberger2016structure}, OpenMVG~\cite{moulon2017openmvg}, or PixelPerfect~\cite{lindenberger2021pixel}.

Here is an outline of the novel view synthesis procedure using NeRFs (\cf (a--c) in Figure~\ref{fig:nerf_pipeline}).
\begin{enumerate}
    \item [i] Send camera rays throughout the image pixels across the scene to produce sampling points.
    \item [ii] Use the MLP(s) to compute local color and density data for those sampling points with corresponding viewing direction.
    \item [iii] Compute the volume rendering to reconstruct the output image by integrating color and density information, throughout.
\end{enumerate}

\subsection{Positional Encoding}
By processing the scene with the standard method we have described so far, experiments show that small displacements in input spatial coordinates to the MLPs $F_\Theta$ may result in sometimes severe outcomes in synthetized images, in particular in high-frequency textured areas. To mitigate this problem, Mildenhall\etal\cite{mildenhall2020nerf} considered positional encoding, which is the mapping of the coordinate inputs to a higher dimensional space using non-linearities prior to passing them to the neural network. This enables better fitting of data that contain high-frequency variations. The encoding function takes the following form:
\begin{equation}
    \gamma(\mathbf{x}) = \left[\sin(\mathbf{x}),\cos(\mathbf{x}),\ldots,\sin(2^{L-1}\mathbf{x}),\cos(2^{L-1}\mathbf{x})\right]
\end{equation}
where $\gamma(\cdot)$ is separately applied to each normalized coordinate value in $\mathbf{x}$ and to the three components of viewing direction unit vector $\mathbf{d}$ where $L$ is the encoding dimensionality parameter (typically $L = 10$ for $\mathbf{x}$ and $L = 4$ for $\mathbf{d}$~\cite{mildenhall2020nerf}).

\subsection{Depth Rendering}
Depth is a valuable source of data for view synthesis and 3D representations. Depth values from a particular pose are calculated in a similar fashion to rendering RGB pixels:
\begin{equation}
    \hat{D}(\mathbf{r}) = \sum_{i=1}^{N}T_i(1-\exp(-\sigma_i\delta_i))t_i 
\end{equation}
where $\hat{D}(\mathbf{r})$ is the expected depth along the camera axis of ray $\mathbf{r}$ and the weighting term is the ray-termination probability of sample $i$ along the ray, defined earlier when rendering color. Following this idea, alternative sources of depth supervision are often utilized to enhance the training and enforce consistency between photometric and geometric constraints. These sources include LiDAR depth~\cite{rematas2022urban}; depth cameras; projected point clouds from Structure from Motion packages~\cite{deng2022depth} or pre-trained depth estimation/completion models~\cite{roessle2022dense}.

Depth loss $\mathcal{L}_{depth}$ is defined in various formulations with the most widely used approach being the MSE between the predicted depth values $\hat{D}(\mathbf{r})$ and the ground truth depth values $D(\mathbf{r})$.
\begin{equation}
    \mathcal{L}_{depth} = \frac{1}{|\mathcal{R}|}\sum_{r\in \mathcal{R}}\left\|\hat{D}(\mathbf{r}) - D(\mathbf{r})\right\|_2^2
\end{equation}
Furthermore, some methods~\cite{niemeyer2022regnerf, xu2022sinnerf} incorporate depth smoothness constraints in addition to the depth loss. This is based on the observation that real-world geometry often exhibits piece-wise smooth characteristics, where flat surfaces are more common than high-frequency structures. To enforce depth smoothness, these methods typically introduce penalties that encourage neighboring pixels of a rendered patch to have similar depth values.
\begin{equation}
    \mathcal{L}_{smooth}(d_i) = e^{-\nabla^2\mathcal{I}(x_i)}(|\partial_{xx}d_i| + |\partial_{xy}d_i| + |\partial_{yy}d_i|)
\end{equation}
where $d_i$ is the depth map, $-\nabla^2\mathcal{I}(x_i)$ refers to the Laplacian of pixel value at location $x_i$.



\begin{figure}[!htbp]
    \centering
    \includegraphics[width=\linewidth]{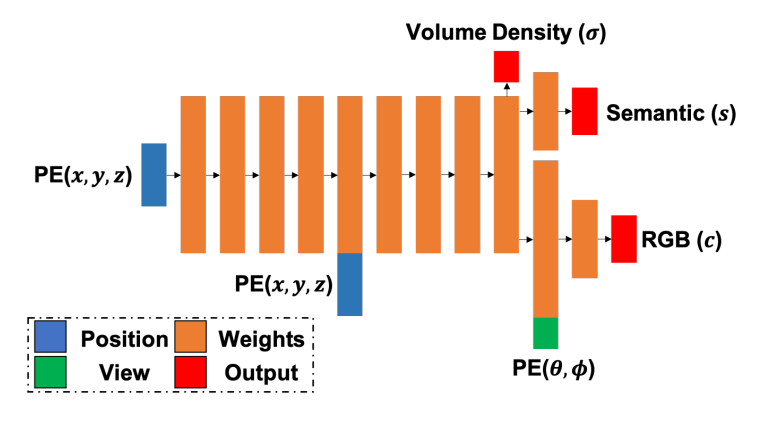}
    \label{fig:semantic_nerf_architecture}
    \caption{\textbf{Semantic NeRFs~\cite{zhi2021place}.} 3D positions $(x, y, z)$ and viewing directions $(\theta, \phi)$ are fed into the network after positional encoding (PE). Volume densities $\sigma$ and semantic logits $\mathbf{s}$ are functions of $(x, y, z)$ while $\mathbf{c}$ additionally depend on $(\theta, \phi)$.}
\end{figure}

\subsection{Empowering NeRFs with Semantic Reasoning}
Research on neural field representations has shown that MLP networks can be trained from scratch for complex scenes by predicting their volumetric density and color supervised solely by a set of RGB images. However, radiance fields only provide low-level representations of geometry and radiance and lack a higher-level (e.g., semantic or object-centric) understanding of the scene. The standard NeRF approach typically suffers from slow training and fails to recover reliable geometry in some cases when the number of input views is sparse and the depth range is infinite~\cite{jain2021putting, fu2022panoptic}. They are also restricted to learning efficient representations of static scenes that encode all objects of the scene and lack the ability to represent complex scenes and the decomposition into individual objects~\cite{ost2021neural} that populate the scene.

\begin{figure*}[!thbp]
    \centering
    \begin{subfigure}[b]{0.4\linewidth}
        \includegraphics[width=\linewidth]{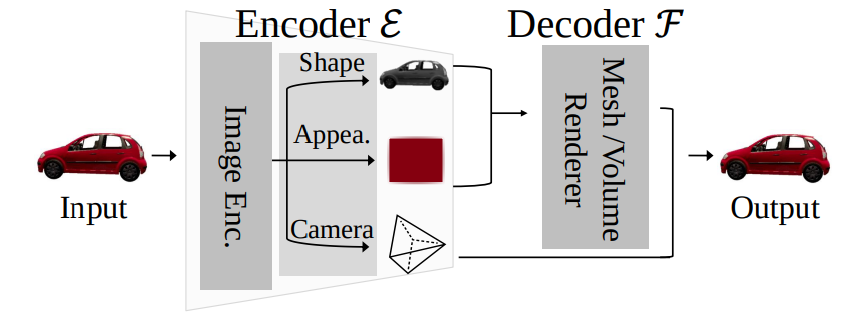}
        \caption{}
    \end{subfigure}
    \begin{subfigure}[b]{0.4\linewidth}
        \includegraphics[width=\linewidth]{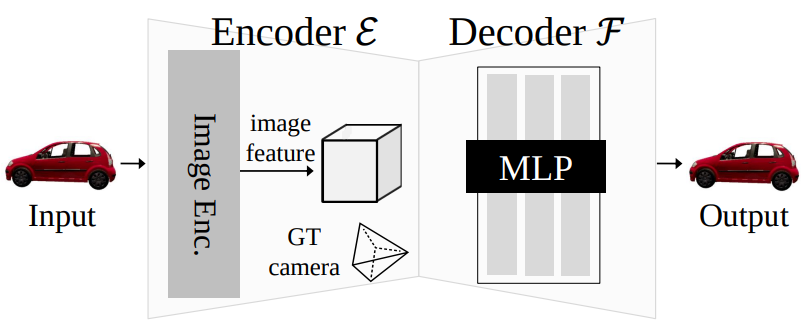}
        \caption{}
    \end{subfigure}
    \begin{subfigure}[b]{0.4\linewidth}
        \includegraphics[width=\linewidth]{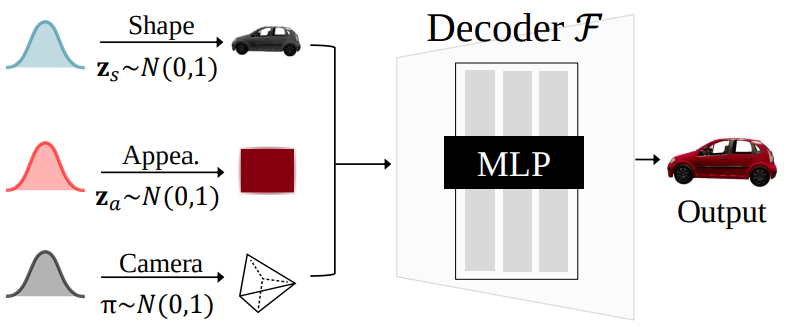}
        \caption{}
    \end{subfigure}
    \begin{subfigure}[b]{0.4\linewidth}
        \includegraphics[width=\linewidth]{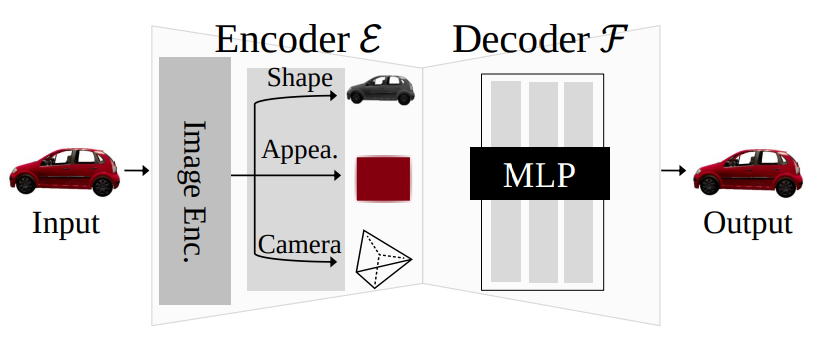}
        \caption{}
    \end{subfigure}
    \caption{Different approaches for conditional 3D representation, which can be effectively used for 3D-aware object manipulation: (a) conditional surface or volumetric representation~\cite{kanazawa2018learning, bhattad2021view}, (b) image-conditional NeRFs~\cite{yu2021pixelnerf, chibane2021stereo, trevithick2021grf, sharma2022neural} that train the feature encoder and NeRF as decoder (c) generative NeRFs~\cite{schwarz2020graf, chan2021pi, xue2022giraffe} that render images from randomly sampled disentangled 3D attributes, and (d) auto-encoding NeRFs~\cite{jang2021codenerf, liu2021editing, kim2022ae} that extract the disentangled 3D latent codes from input and renders images from these attributes.}
    \label{fig:intuition_conditional_nerf}
\end{figure*}

\subsubsection{Semantic Radiance Fields}
Semantic labels can also be formalized as an inherently view-invariant function that maps only a world coordinate $\mathbf{x}$ to a distribution over semantic labels via pre-softmax semantic logits $\mathbf{s}(\mathbf{x})$, while for instance, it is a one-hot encoding of the object instance identifier. This was done by appending additional branches before injecting the viewing direction $\mathbf{d}$ into the rendering function:
\begin{equation}
    \hat{S}(\mathbf{r}) = \sum_{i=1}^{N}T_i(1-\exp(-\sigma_i\delta_i))\mathbf{s}_i 
\end{equation}
Semantic logits can then be transformed into multi-class probabilities through a softmax normalization layer. During inference, the semantic label is determined as the class of the maximum probability in $\hat{S}(\mathbf{r})$.

The semantic loss $\mathcal{L}_{sem}$ is usually chosen as a multi-class cross-entropy loss to encourage the rendered semantic/instance labels to be consistent with the provided labels, whether these are ground-truth, noisy, or partial observations:
\begin{equation}
    \mathcal{L}_{sem}^{2D} = -\frac{1}{|\mathcal{R}|}\sum_{r\in \mathcal{R}}\left[\sum_{l=1}^{L}p^l(\mathbf{r})\log\hat{p}^l(\mathbf{r})\right]
\end{equation}
where $1 \leq l \leq L$ denotes the class index, $p^l(\mathbf{r})$, $\hat{p}^l(\mathbf{r})$ are the multi-class semantic probability (by forwarding the logits into a softmax normalization layer) of the camera ray $\mathbf{r}$ at class $l$ of the ground truth and predicted map.

\subsubsection{Prior Learning and Conditional NeRFs}
A conditional neural field introduces the ability to alter the characteristics of a radiance field through the manipulation of latent variables $\mathbf{z}$. These latent variables can encompass diverse aspects, ranging from random samples drawn from any distribution to geometric/semantic attributes such as shape, type, size, color, and more. Alternatively, they could be derived from the encoding of other data types, including embedded text or audio data. 
Instance-specific details can be encoded within the conditioning latent variable z, whereas information shared across instances is stored in the parameters of the neural field. When these latent variables are mapped to a semantic or smoothly varying space, it allows for their interpolation or editing.
%

The conditioning latent code $\mathbf{z} = \mathcal{E}(I)$ is generated by an encoder or embedding mechanism $\mathcal{E}$ usually implemented as a neural network (as shown in Figure \ref{fig:intuition_conditional_nerf}). The parameters within $\mathcal{E}$ are capable of encoding prior knowledge, which can be learned from pre-training on datasets or through auxiliary tasks.
%
The decoder is the neural field that is conditioned by the latent code:
\begin{equation}
    (\mathbf{c}, \sigma) = F_\Theta(\mathbf{x}, \mathbf{d}, \mathbf{z})
\end{equation}
This adaptation can be achieved by conditioning the field on latent variables $\mathbf{z}$ that encapsulate specific higher-level, semantic characteristics from the scene. When these latent variables are edited, the corresponding neural field can be modified accordingly. 

\subsubsection{High-level Feature Consistency}
While the field of 3D scenes has unique challenges and complexities, the image domain stands out with its abundance of extensive, high-quality datasets and a wealth of established techniques for effective feature extraction. The semantic richness captured in image feature spaces can be harnessed to establish correspondences and enhance understanding through text, image queries, or clustering. Although there exists pixel-wise misalignment between the views, it is observed that the extracted representation of pre-trained deep neural networks as feature extractors is robust to this misalignment and provides supervision at the semantic level~\cite{jain2021putting,xu2022sinnerf,mirzaei2023spin}. 
Intuitively, this occurs naturally because the content and style of the two views are alike, allowing a deep network to learn a representation that remains consistent across them.
Perceptual Loss $\mathcal{L}_{feat}$, also known as feature loss or content loss, is a measure of the discrepancy between the high-level features of the predicted image and the ground truth image, both extracted from the pre-trained network:
\begin{equation}
    \mathcal{L}_{feat} = \frac{1}{N}\sum_{i=1}^{N}\left\|\Phi(\hat{I}_i) - \Phi(I_i)\right\|_2^2
\end{equation}
where $\Phi(\cdot)$ refers to the extracted features/tokens. $I$ and $\hat{I}$ are patches or images from the reference view and rendered view, respectively. This commonly appears in tasks where preserving high-level global features is important.


It is shown~\cite{kobayashi2022decomposing, tschernezki2022neural} that the knowledge distilled from the teacher model aligns with the scene's geometry, thereby enhancing feature quality across viewpoints and occlusion awareness; and the infusion of features pre-trained on diverse external datasets, bringing a broader open-world perspective to the 3D representation without collecting annotations for them.

\begin{figure*}[thbp]
\centering
\begin{tikzpicture}[]
    \def\w{\textwidth}
    \newcount\yearOne; \yearOne=2020
    \def\tl{0.1} 
    \def\yearGap{4.8}

    \def\yearArrowLabel(#1,#2,#3,#4,#5){
        \def\xy{{(#1-\yearOne-1)*\yearGap+1}}; \pgfmathparse{int(#2)};
        \ifnum \pgfmathresult<0
          \def\yyp{{(\tl*(#2))}}; \def\yyw{{(\yyp-\tl*#3)}}
          \draw[<-,thick,#5,align=center] (\xy,\yyp) -- (\xy,\yyw) node[below,#5] at (\xy,\yyw) {#4};
        \else
          \def\yyp{{(\tl*(#2)}}; \def\yyw{{(\yyp+\tl*#3)}}
          \draw[<-,thick,#5,align=center] (\xy,\yyp) -- (\xy,\yyw) node[above,#5] at (\xy,\yyw) {#4};
        \fi}

    \def\yearDotLabel(#1,#2,#3,#4){
        \def\x{{(#1-\yearOne-1)*\yearGap+1}}; 
        \fill[#4,radius=2pt] (\x,0) circle node[#3,#4]{#2};
    }
    
    \draw[->,line width=0.5mm] (0, 0) -- (\w,0);
    \foreach \tick in {0,1,2,3}{
        \def\x{1+\tick*\yearGap}
        \def\year{\the\numexpr \yearOne+\tick+1}
        \draw[thick] (\x,\tl) -- (\x,-\tl) node[below] {\year};
    }

    \yearArrowLabel(2020.8, -1, 5, NeRF\\baseline, black)
    \yearArrowLabel(2021.0, 1, 5, NeRF-W, red)
    \yearArrowLabel(2021.3, 1, 5, DietNeRF, blue)
    \yearArrowLabel(2021.3, 12, 5, NSG\\NSFF, red)
    \yearArrowLabel(2021.3, -1, 15, GRAF\\pi-GAN, teal)
    \yearArrowLabel(2021.4, -1, 5, pixelNeRF, blue)
    \yearArrowLabel(2021.5, 1, 24, EditNeRF, teal)
    \yearArrowLabel(2021.7, 1, 5, MVSNeRF, blue)
    \yearArrowLabel(2021.7, 12, 5, S-NeRF, violet)
    \yearArrowLabel(2021.8, -1, 12, Object-NeRF, red)
    \yearArrowLabel(2022.2, 1, 5, CLIP-NeRF, orange)
    \yearArrowLabel(2022.3, -1, 15, Control-NeRF, teal)
    \yearArrowLabel(2022.4, -1, 5, PNF, red)
    \yearArrowLabel(2022.4, 1, 12, DreamField, orange)
    \yearArrowLabel(2022.5, -1, 20, ShaPO, magenta)
    \yearArrowLabel(2022.7, 1, 5, SinNeRF, blue)
    \yearArrowLabel(2022.7, -1, 15, GRAM, teal)
    \yearArrowLabel(2022.9, -1, 12, Latent-NeRF, orange)
    \yearArrowLabel(2022.8, -1, 5, DFF, violet)
    \yearArrowLabel(2023.0, 1, 10, Panoptic Lift, violet)
    \yearArrowLabel(2023.2, 1, 5, Zero-1-to-3, blue)
    \yearArrowLabel(2023.2, 12, 5, NeRF-RPN, magenta)
    \yearArrowLabel(2023.3, -1, 5, LeRF, orange)
        \yearArrowLabel(2023.5, -1, 12, NeRF-Det, magenta)
    \yearArrowLabel(2023.5, 1, 10, Instruct-N2N, orange)
    \yearArrowLabel(2023.8, -1, 5, NeO360, blue)
    \yearArrowLabel(2024.0, 1, 5, 3D-OVS, violet)

    \end{tikzpicture}
    \caption{Chronological overview of the most relevant semantically-aware NeRFs spanning all 6 categories covered by our study: \textcolor{blue}{3D geometry enhancement}, \textcolor{violet}{segmentation}, \textcolor{teal}{editable NeRFs}, \textcolor{magenta}{object detection and 6D pose}, \textcolor{red}{holistic decomposition}, \textcolor{orange}{NeRFs and language}.}
    \label{fig:timeline}
\end{figure*}

\section{Semantically-aware NeRFs for Visual Scene Understanding}
\label{sec:review}
In this section, we review the most prominent NeRF-based approaches and strategies that either use semantic-level reasoning as leverage to enhance 3D geometry or that aim to achieve a higher level of scene understanding through either of the tasks and applications considered in our considered taxonomy (Fig.~\ref{fig:taxonomy} and Fig.~\ref{fig:timeline}).

\subsection{3D Geometry Enhancement}
\label{sec:improve_3d}
In this category, several notable approaches incorporate semantic reasoning to improve performance in Novel View Synthesis (NVS), to make up for limited amounts of input views, to generalize to unseen environments, or to address 3D surface reconstruction.

\subsubsection{One-shot/Few-shot NeRFs}

\begin{figure*}[!thbp]
    \centering
    \includegraphics[width=\linewidth]{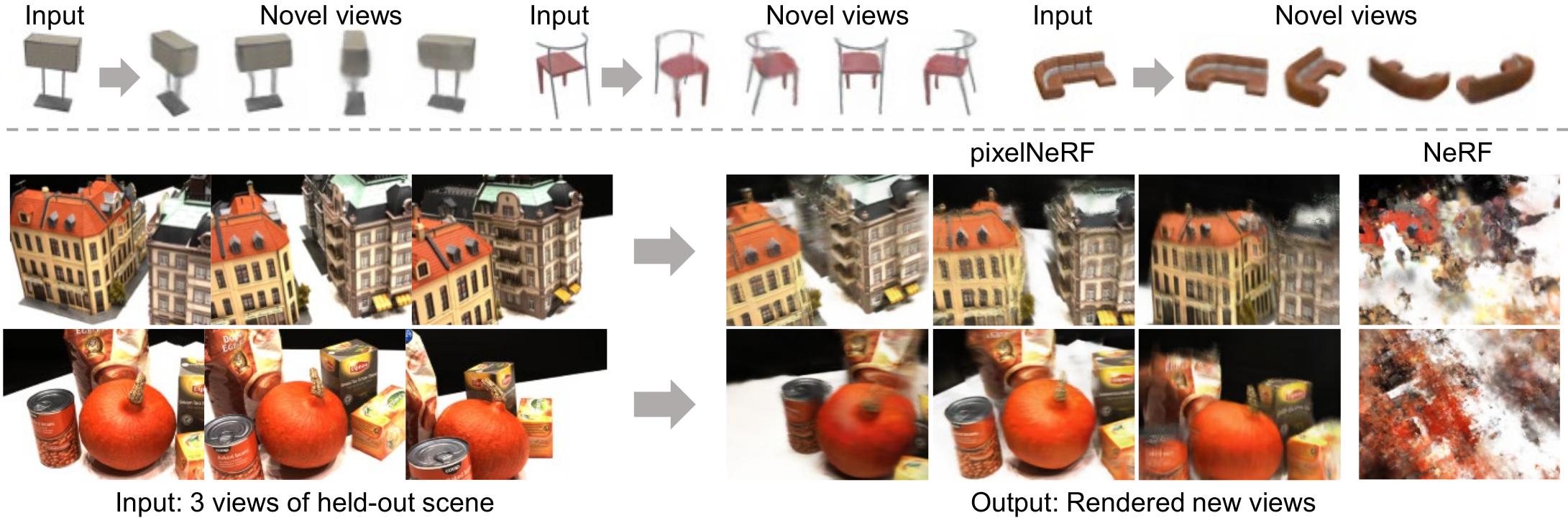}
    \caption{\textbf{Generalization capabilities of PixelNeRF~\cite{yu2021pixelnerf}} -- The method can be trained from multiple datasets in order to synthesize plausible novel views from very few input views, without test-time optimization. In contrast, standard NeRF can not generalize to previously unseen environments.}
    \label{fig:pixel_nerf_application}
\end{figure*}

\textbf{PixelNeRF}~\cite{yu2021pixelnerf} (Fig.~\ref{fig:pixel_nerf_application}) and \textbf{S-RF}~\cite{chibane2021stereo} use image-level CNN features, whereas \textbf{MVSNeRF}~\cite{chen2021mvsnerf} builds a 3D cost volume via image warping which is then processed by a 3D CNN. This fully convolutional strategy allows the network to be trained across multiple scenes to learn scene-level priors and, thus, generalize to unseen environments and object categories. 
Building on this concept, \textbf{MINE}~\cite{li2021mine}, \textbf{Behind the Scenes}~\cite{wimbauer2023behind}, and \textbf{SceneRF}~\cite{cao2023scenerf} reduce the scene representation complexity by leveraging monocular depth estimation and redefine feature extraction and rays and color sampling accordingly to account for the self-supervised depth network.
\textbf{DietNeRF}~\cite{jain2021putting} and \textbf{SinNeRF}~\cite{xu2022sinnerf} match high-level and global semantic attributes to semantic pseudo-labels with texture guidance across different views, allowing us to supervise the training process from random poses. This improves the perceptual quality of NVS in the few-shot setting in particular.

Single-view (\ie one-shot) reconstruction can also be formulated as a conditioned 3D generation problem for a single-image NVS task without explicit 3D supervision. \textbf{RealFusion}~\cite{melas2023realfusion} and \textbf{Zero-1-to-3}~\cite{liu2023zero} extract a neural field from the original image input and a internet-level pre-trained diffusion models, thus achieving a comprehensive reconstruction of the object from unseen viewpoints, or in a prompt-constrained zero-shot setting. This process captures both appearance and geometry. Additionally, image-level text embeddings can be extracted through textual inversion, which captures additional high-level visual cues. However, such a strategy yields substantially ambiguous representations in unobserved areas, and they are mostly object-centric assuming a plain background. 

\textbf{NeRDi}~\cite{deng2023nerdi} also uses diffusion priors trained on large image datasets. It utilizes a two-section semantic guidance to refine the general prior knowledge conditioned on the input image. This ensures that synthesized novel views are both semantically and visually consistent. Despite the training of the model being carried out on a synthetic dataset, it shows robust zero-shot generalization capabilities. It effectively extends to both out-of-distribution datasets and real-world, in-the-wild images.
\textbf{SegNeRF}~\cite{zarzar2022segnerf} and \textbf{S4C}~\cite{hayler2023s4c} address generalization and learn a semantic field, performing reconstruction and segmentation in a self-supervised fashion from a single view, while also allowing for semantic object/scene completion. 
\textbf{Neural groundplans}~\cite{sharma2022neural} conditions a self-supervized NeRF on ground-aligned 2D feature grids trained from multi-view videos. \textbf{NeO 360}~\cite{irshad2023neo} leverages a hybrid conditional triplanar representation which combines the strengths of voxel and bird's eye view (BEV) representation. These hybrid discrete-continuous representations allow to learn from a large collection of 360 unbounded  scenes while addressing different downstream tasks including NVS, object localization, and scene editing from as few as a single image during inference.

\subsubsection{Surface Reconstruction}
The piecewise planarity assusmption, \ie which assumes that a given scene can be mostly explained by piecewise planar surfaces has been a stable prior in the traditional 3D reconstruction literature and has also proven effective in the context of implicit neural representations. Guo\etal~\cite{guo2022neural} formulate geometric constraints of floors and walls within the normal loss function adhering to the Manhattan World Assumption~\cite{coughlan2000manhattan}, assuming three mutually orthogonal surface orientations. These regions were obtained by a 2D semantic segmentation networks. To address inaccurate segmentations, they encode the semantics of 3D points with another MLP that jointly optimizes the scene geometry and semantics.
\textbf{PlaNeRF}~\cite{wang2023planerf} also performs a planar regularization based on Singular Value Decomposition (SVD). This improves the underlying geometry in that correspond to image regions with low texture, without any additional geometric prior.
\textbf{S3PRecon}~\cite{ye2023self} introduces an iterative training scheme for grouping pixels and optimizing the reconstruction network via a superplane constraint. This in particular yields better performance than using explicit 3D plane supervision which is costly to obtain.

\textbf{SS-NeRF}~\cite{zhang2023beyond} and \textbf{MuvieNeRF}~\cite{zheng2023multi} are versatile multi-task frameworks. They can render images from novel viewpoints and manage various scene properties such as appearance, geometry, and semantic segmentation. Both utilize a shared scene encoding network that allows for cross-view and cross-task attention modules to ensure view consistency. They also examine the relationships among different scene properties to enhance performance. This approach highlights the potential of multi-task learning and knowledge transfer within a synthesis paradigm in benefiting from the mutually-informative relationships between different tasks and properties, \eg semantic labels, surface normal, shading, keypoints, and edges.

\subsection{Segmentation}\label{sec:segmentation}
The most common approach for scene understanding typically focuses on 2D reasoning in image space, using classic image-to-image architectures that are trained on extensive sets of semantically annotated images. Although these techniques are easy to implement, they generate only pixel-by-pixel annotations and mostly overlook the underlying 3D structure of the scene. In contrast, our objective is to use a set of RGB images with established poses to produce a 3D semantic/instance field. This involves devising a function that assigns probability distributions over semantic and/or instance-level categories to specific 3D positions and viewpoints.


\subsubsection{Semantic, Instance, and Panoptic Segmentation}
\textbf{NeSF}~\cite{vora2022nesf} uses a pre-trained NeRF to generate a volumetric density grid. Following this, a 3D UNet is used to produce a feature grid that maintains the same spatial resolution. This process enables high-level reasoning within the 3D space. Semantic maps are generated through the application of the volumetric rendering equation, using camera poses on the semantic field. Consequently, NeSF is trained comprehensively on various scenes, eliminating the need for segmentation input when making inferences about new scenes.
%

\textbf{Semantic-NeRF}~\cite{zhi2021place} is a groundbreaking work that extends NeRF to include both semantics along with appearance and geometry. By adding semantic class predictions to radiance and density within a scene-specific implicit MLP model, it can ensure multi-view consistency between semantic labels. Consequently, the experiments demonstrate its ability to perform multi-view semantic label fusion in various scenarios: pixel-wise label noise, region-wise label noise, low-resolution dense or sparse labeling, partial labeling, and using the output from an imperfect segmentation model.
%
%
%
In this respect, several studies leverage 3D geometry together with semantic predictions to resolve label uncertainties. For example, 
\textbf{Panoptic NeRF}~\cite{fu2022panoptic, fu2023panopticnerf} introduces an optimization process guided by semantics to enhance the underlying geometry. This technique uses a dual of semantic fields: a fixed semantic field which focuses on guiding the underlying density, defined by 3D bounding primitives, and a learned semantic field designed to capture the semantic distribution. 

Another work, \textbf{Semantic Ray}~\cite{liu2023semantic}, fully exploits semantic information along the ray direction from its multi-view re-projections. The authors tackle the limitations of prior methods that depend on positional encoding and scene-specific models for semantic learning. Unlike these approaches, they harness insights from multiple views using a new module called Cross-Reprojection Attention. This module efficiently captures contextual information along the reprojected ray paths, enriching the understanding from various views.
%
%

\textbf{JacobiNeRF}~\cite{xu2023jacobinerf} introduces a regularization of learning processes to align the Jacobians of highly correlated entities, effectively maximizing their mutual information amid random perturbations in the scene. This approach of mutual information modeling is key in configuring NeRF to perform sparse label propagation for semantic and instance segmentation. For a given target view of a scene that is unlabeled, one can produce labels by selecting the argmax of the perturbation responses from the source view annotations.

Liu\etal ~\cite{liu2023unsupervised} propose the training of a Semantic-NeRF network for each scene by fusing the predictions of a segmentation model and using the view-consistent rendered semantic labels as pseudo-labels for model adaptation. Their method simultaneously trains the frame-level semantic network and the scene-level NeRF, ensuring that the semantic forecasts and NeRF renderings are in alignment. This transfer strategy not only boosts the performance of both models but also reflects a real-world deployment scenario that accounts for the covariate shift across different scenes and the possibility of revisiting previously observed scenes.

%
%
Traditional methods depend on accurately labeled ground truth data to train models for object-compositional scene representations. It's important to recognize that these manual annotations are designed to be 3D consistent, ensuring that identifiers for specific objects remain constant across different viewpoints. However, a major challenge presents itself when using pseudo-labels generated by off-the-shelf networks. These labels, inferred from individual views, often fail to maintain the 3D alignment of instance indices, leading to inconsistency.
Several studies have focused on addressing the discrepancies and maintaining consistency across different viewpoints within the same scene, particularly when employing an off-the-shelf 2D panoptic segmentation network. 
These efforts strive to preserve the object instance identities from machine-generated panoptic labels within the implicit 3D volumetric representation.
For instance, \textbf{Panoptic Lifting}~\cite{siddiqui2023panoptic} assigns 3D surrogate identifiers to machine-generated instances by solving linear assignment problems, using these associations to guide the training of the instance field through an NCE loss.

\textbf{Contrastive Lift}~\cite{bhalgat2023contrastive} introduces a change to the labeling process by using a low-dimensional Euclidean space, which simplifies the model by reducing the dimensions needed to calculate pairwise distances. 
This slow-fast clustering objective function is scalable and suitable if there are a large number of objects (up to 500 per scene).
On the other hand, \textbf{PCFF}~\cite{cheng2023panoptic} proposes an Instance Quadruplet loss which leads to a discriminating feature space for the scene decomposition at instance levels. 
The model is further refined with strategies that are added to the architecture, like semantic-appearance hierarchical learning and semantic-guided regional refinement.
Finally, \textbf{Instance-NeRF}~\cite{liu2023instance} seeks to match 3D object masks projected from a proposal-based NeRF-RCNN with inconsistent segmentation maps in image space, thereby refining the initial instance segmentation results.

\begin{figure*}[!htbp]
    \centering
    \includegraphics[width=\linewidth]{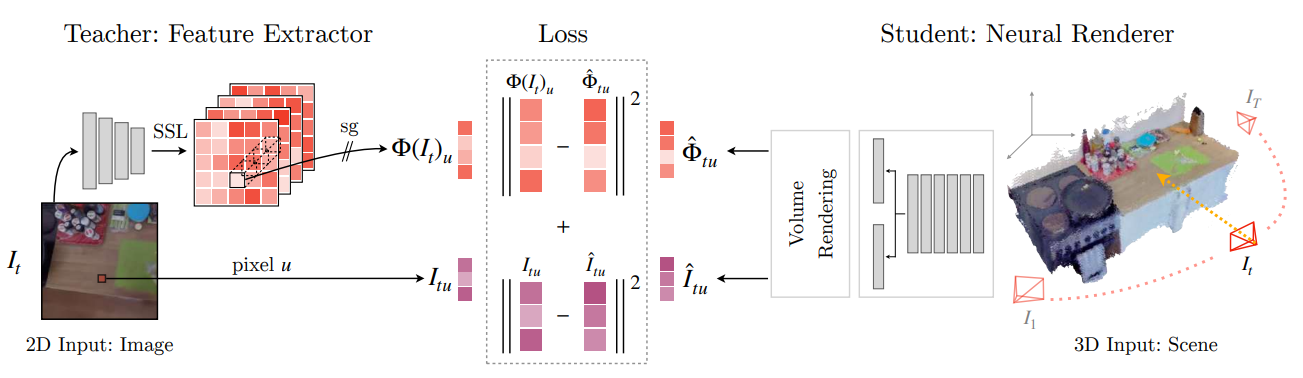}
    \caption{\textbf{N3F paradigm~\cite{tschernezki2022neural}} - A student-teacher framework distills predicted 2D features from images into a 3D student NeRF-like model. The student optimizes image and feature reconstruction, while the teacher remains untrained. The resulting representation can operate in 2D or 3D contexts.}
    \label{fig:N3F_architecture}
\end{figure*}

\subsubsection{Pre-Semantic Segmentation}
\begin{figure*}[!htbp]
    \centering
    \includegraphics[width=\linewidth]{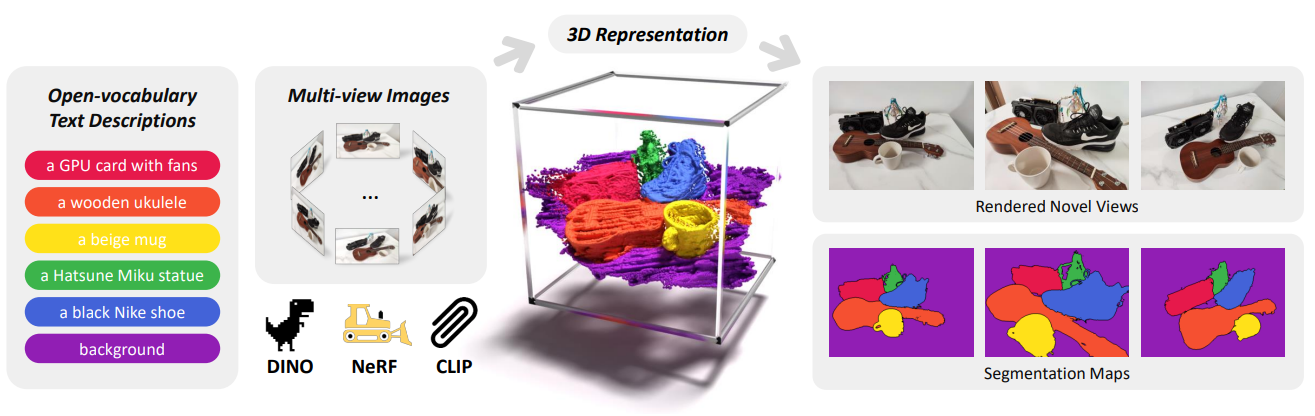}

\caption{The method developed by Liu\etal \cite{liu2023weakly} merges multimodal open-vocabulary knowledge from CLIP~\cite{radford2021learning} with the object reasoning capabilities of DINO~\cite{caron2021emerging} resulting in accurate delineation of 3D objects without relying on segmentation annotations during training. This showcases the model’s ability to render novel views with corresponding segmentation maps.}
    \label{fig:weekly_seg_application}
\end{figure*}

\textbf{DFF}~\cite{kobayashi2022decomposing}, \textbf{N3F}~\cite{tschernezki2022neural} and \textbf{FeatureNeRF}~\cite{ye2023featurenerf} adopt a 2D-teacher-3D-student framework. In this setup, pre-trained 2D image feature extractors like LSeg~\cite{li2022languagedriven}, SAM~\cite{kirillov2023segment}, and DINO~\cite{caron2021emerging} act as `teachers' that guide the learning process of a NeRF `student' network. The loss function in this context is designed by imposing penalties on the discrepancies between rendered features and the outputs generated by the feature descriptor. These methods pave the way for applications in language-guided editing, 3D spatial rearrangements, and targeted scene removal.
%
%
\textbf{NeRF-SOS}~\cite{fan2022nerf} integrates a self-supervised pre-trained framework to generate feature tensors from color patches rendered by the model. This approach then uses these features to create volumes for appearance-segmentation, applying contrastive losses to correlate both appearance-segmentation and geometry-segmentation. During inference, the model perform a clustering process on the rendered feature field to produce segmentation masks. Similarly, \textbf{3D-OVS}~\cite{liu2023weakly} demonstrates that aligning the class relevancy distribution with these pre-trained foundation models in a weakly supervised way can achieve precise, annotation-free segmentation, as shown in Figure \ref{fig:weekly_seg_application}. \textbf{Feature-Realistic Fusion}~\cite{mazur2023feature} fuses general features learned from EfficientNet into NeRF representation. With a SLAM backend, this system operates incrementally in real-time, effectively managing the exploration of new, unobserved regions of a scene.
%

\textbf{RFP}~\cite{liu2022unsupervised} introduces an innovative propagation method that uses a bidirectional photometric loss. This approach allows for unsupervised partitioning of a scene into distinct, salient regions that correspond to individual object instances, effectively performing object segmentation within the scene. \textbf{IntrinsicNeRF}~\cite{ye2023intrinsicnerf} goes further by producing outputs like reflectance, shading, and residual terms. The model is trained using unsupervised prior and reflectance clustering as constraints in the loss function. These terms are particularly useful for real-time augmented applications such as recoloring, illumination variations, and, importantly, semantic segmentation.

\textbf{SNeRL}~\cite{shim2023snerl} integrates NeRF with semantic and distilled feature fields specifically for reinforcement learning applications. It employs a NeRF-based autoencoder, trained to act as a feature extractor, for fine-tuning in multi-view reinforcement learning (RL) tasks. This method has shown to outperform current representation learning techniques in both model-free and model-based RL algorithms across various 3D environments.
%
%

\subsubsection{Interactive Segmentation}
For a practical scene-annotation tool, simple user annotations like sparse clicks can be extended and propagated to achieve dense and accurate labeling of the scene. This process allows for the creation of complete and accurate 2D semantic labels with minimal in-scene annotations specific to the scene. 
\textbf{iLabel}~\cite{zhi2021ilabel} takes this concept further by integrating semantic label-propagation into an online, interactive 3D scene-capturing system, enabling segmentation of coherent 3D entities with minimal user click annotations. The authors also introduced a novel hierarchical semantic representation using a binary tree, facilitating the prediction of semantics at different levels.
\textbf{Baking in the Feature}~\cite{blomqvist2023baking} and \textbf{ISRF}~\cite{goel2023interactive} merge distilled features with a bilateral search in a unified spatial-semantic space for an interactive segment user interface. 
\textbf{NVOS}~\cite{ren2022neural} trains a 3D segmentation network to classify each voxel as foreground or background, using partial user scribbles as supervision. This is followed by applying the learned classifier and further refining the segmentation with a 3D graph-cut, leveraging the 3D distance field of the scribble.
Other methods~\cite{cen2023segment, wei2023noc} aim to generalize the Segment Anything Model (SAM)~\cite{kirillov2023segment} for 3D object extraction. These alternate between mask inverse rendering and cross-view self-prompting across different views to iteratively complete the 3D object mask from a single view. Users can annotate frames in an RGB video sequence with brush strokes, while the system concurrently fits a model to the scene and annotations.
These strategies surpass the labeling accuracy of conventional pre-trained semantic segmentation methods. \textbf{SGISRF}~\cite{tang2023scene} takes this even further, requiring fewer user interactions for interactive segmentation by using Cross-Dimension Guidance Propagation and Concealment-Revealed Learning schemes.
\begin{figure}[!htbp]
    \centering
    \includegraphics[width=\linewidth]{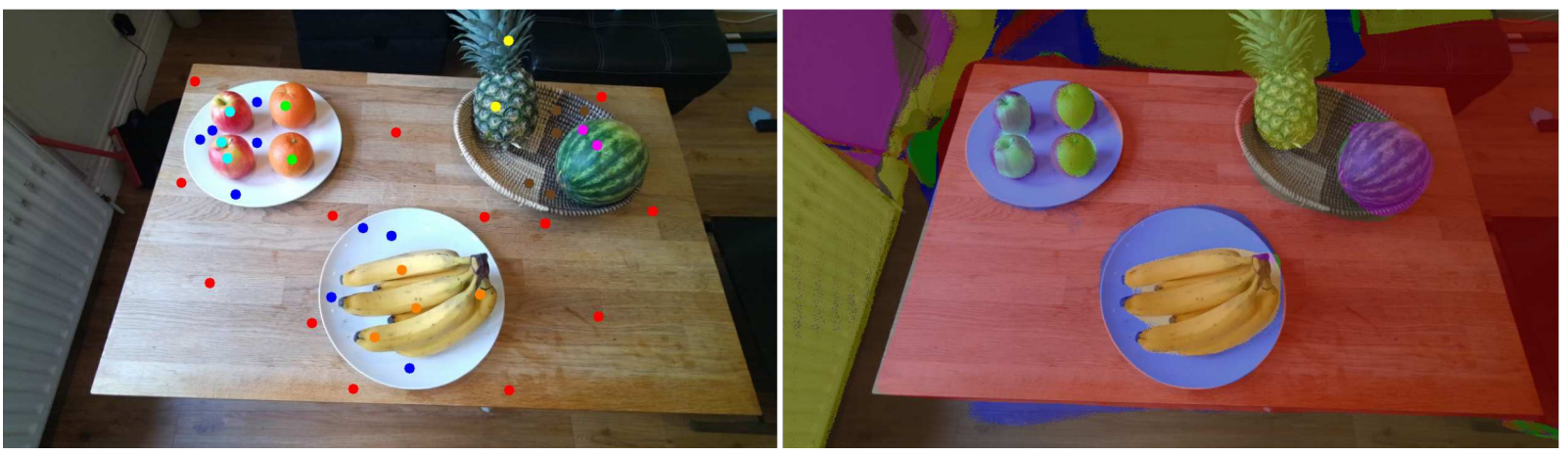}
\caption{iLabel~\cite{zhi2021ilabel} demonstrates the capability to create high-quality segmentations of various entities within a scene using only a minimal number of user-provided clicks.}
%
    \label{fig:ilabel_application}
\end{figure}
Another key interactive feature in 3D scene manipulation is the removal of undesired objects in a way that the resulting area blends seamlessly and logically with its surroundings, a process commonly known as image inpainting. This technique starts with a pre-trained NeRF model and its associated image dataset. In the first stage, known as mask generation, an initial mask is created from a single-view annotation using one-shot segmentation methods like Mask R-CNN~\cite{he2017mask}, SAM~\cite{kirillov2023segment}, or GLIP~\cite{li2022grounded, wang2023inpaintnerf360}. Following this, \textbf{NeRF-In}~\cite{liu2022nerf} uses an inpainting network~\cite{cao2021learning, suvorov2022resolution} to produce a guiding image and a depth image, based on the user-selected area to be removed. This process updates the NeRF model by 
optimizing both the color-guiding and depth-guiding losses.
While NeRF-In doesn't fully resolve the 3D inconsistencies in the output of the inpainters, and only minimizes the number of views used, there are proposals to overcome blurring and ensure consistency between views. These include using a relaxation approach based on perceptual loss~\cite{mirzaei2023spin}, applying bilateral solvers, and incorporating estimated depth to introduce view-dependent effects in the inpainted regions~\cite{mirzaei2023reference}. Another technique involves selectively excluding views using an uncertainty mechanism and a pixel-wise loss~\cite{weder2023removing}. Weder \etal~\cite{weder2023removing}'s method updates the set of images used for optimization iteratively, based on confidence scores, to maintain consistency during the inpainting process. This enables the generation of realistic novel views of the scene without the removed objects.
\subsection{Editable NeRFs}\label{sec:edit_nerf}
\subsubsection{Conditional NeRFs}
\textbf{CodeNeRF}~\cite{jang2021codenerf} implements the learning of separate embeddings, whereas \textbf{EditNeRF}~\cite{liu2021editing} incorporates a shared shape branch within the conditional radiance field, aiming to better reconstruct shape instances. Both approaches encourage the network to develop a common representation across different object instances, resulting in enhanced shape editing and consistency. \textbf{ShaRF}~\cite{rematas2021sharf} employs a shape network that maps the shape latent code into a 3D shape in the form of a voxel grid. The NeRF network then conditions on two additional factors: the occupancy value estimated from the voxel grid and the appearance latent code that dictates the object’s appearance. 
\textbf{AE-NeRF}~\cite{kim2022ae} introduces two specific losses - global-local attribute consistency loss and swapped-attribute classification loss - to enhance disentanglement capabilities. Furthermore, this conditional model benefits from a GAN-based stage-wise training approach, significantly elevating its performance.
%
%
\textbf{AutoRF}~\cite{muller2022autorf} and \textbf{Car-NeRF}~\cite{liu2023car} develop an object-level radiance field specifically for cars, effectively disentangling shape and appearance within their image encoders. For each car instance, they use a panoptic segmentation mask and a 3D bounding box, which describe the pose and size of the object. These models transform each ray from the camera space into the Normalized Object Coordinate Space (NOCS), creating an object-centric ray that allows for the generation of high-quality car images from any single-view input. Thanks to their ability to perform shape and color edits within the network layers, these models facilitate a hybrid network update strategy. This approach enables the formulation of optimization problems for color and shape editing that meet specific user requirements while maintaining the integrity of the original object's structure. Such features are key in preserving the overall visual coherence of the edited object and reduce the number of images needed during testing.

\subsubsection{Generative NeRFs}
\begin{figure*}[!htbp]
    \centering
    \begin{subfigure}[b]{0.62\linewidth}
        \includegraphics[width=\linewidth]{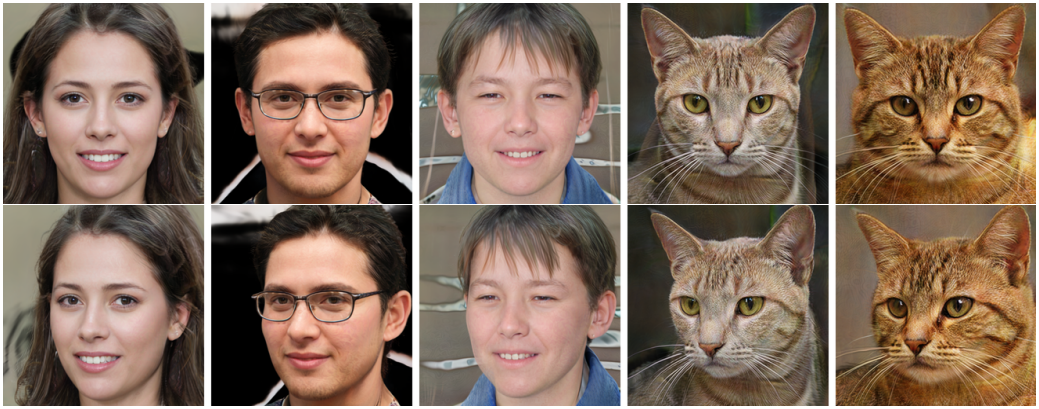}
        \caption{}
        \label{fig:gram_application}
    \end{subfigure}
    \begin{subfigure}[b]{0.36\linewidth}
        \includegraphics[width=\linewidth]{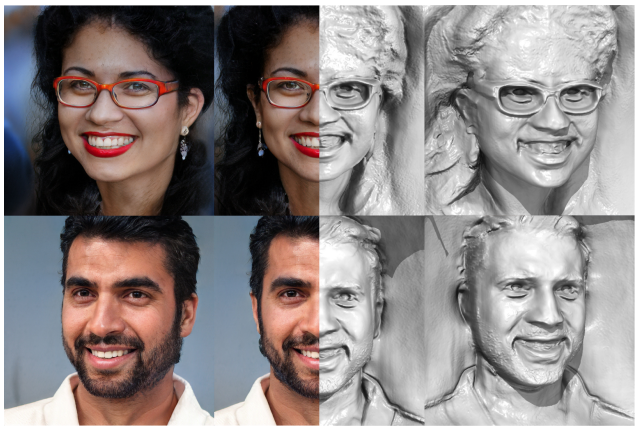}
        \caption{}
        \label{fig:geometry_3d_gan}
    \end{subfigure}
\caption{Illustration of view-controllable images generated by GAN-based NeRFs, showcasing the high quality and 3D consistency of the output. The left set (a) features images from the GRAM~\cite{deng2022gram} study, while the right set (b) includes images from the work of Chan et al.~\cite{chan2022efficient}.}
\end{figure*}
Recent advancements in NeRF-based generative models, including VAEs, GANs, and diffusion models, have significantly progressed in creating 3D-aware generators. These models have the capability to disentangle the underlying 3D aspects of the objects they represent, enabling precise manipulation of camera poses while still producing high-fidelity object renderings. Additionally, these models are designed to generate view-consistent and varied images that accurately reflect specified conditions. The versatility of these models is further enhanced by their ability to incorporate a range of user-defined conditions, such as text and images, into their generation process.
%
%
\textbf{GRAF}~\cite{schwarz2020graf} and \textbf{pi-GAN}~\cite{chan2021pi} introduce a generative model that employs implicit radiance fields for the synthesis of novel scenes. These models are trained on unposed images and focus on simple objects. Building on GRAF, \textbf{GIRAFFE}~\cite{niemeyer2021giraffe} enhances this approach by representing scenes through compositional generative neural feature fields. This advancement allows for the disentanglement of individual objects' shapes and appearances from their backgrounds without the need for explicit supervision. As a result, users are provided with greater flexibility to compose more complex scenes.
%
%
While this method is less demanding on memory when scaling up to higher resolutions compared to voxel-based techniques, it still requires considerable computational power to train and render images at high resolutions. In response, \textbf{StyleNeRF}~\cite{gu2021stylenerf}, \textbf{GIRAFFE HD}~\cite{xue2022giraffe}, and work by Chan \etal~\cite{chan2022efficient} all aim to retain the 3D controllability characteristic of GIRAFFE while generating images of much higher quality and resolution (exceeding $512 \times 512$), using the architecture of StyleGAN2~\cite{karras2020analyzing}. Following in these footsteps, \textbf{UrbanGIRAFFE}~\cite{yang2023urbangiraffe} extends this concept by using coarse panoptic priors in the form of semantic voxel grids and object layouts. This approach enhances controllability even more, particularly for substantial changes in camera viewpoint and semantic layouts.
%

%
Reconstructing a dynamic human face poses unique challenges due to the complexity of facial geometry and the varying appearances caused by diverse expressions. Facial expressions, involving a mix of local deformations, can be represented through controllable attributes defined as latent variables. These attributes can be flexibly applied to different types of conditions such as landmarks~\cite{rebain2022lolnerf}, sketches, low-resolution images, and text~\cite{jo2023cg} as input conditions.
Methods like \textbf{CoNeRF}~\cite{kania2022conerf} and \textbf{FaceCLIPNeRF}~\cite{hwang2023faceclipnerf}, which build upon HyperNeRF~\cite{park2021hypernerf}, are capable of being trained on dynamic scenes to control facial deformations using only sparse input views. Users can manipulate facial attributes effectively by providing simple expression codes~\cite{gafni2021dynamic, zhang2022fdnerf}, mask annotations of facial regions~\cite{sun2022fenerf, kania2022conerf} (such as eyes being open/closed or mouths smiling/frowning), or textual descriptions~\cite{hwang2023faceclipnerf} (like ``happy", ``surprised", ``fearful", ``angry", and ``sad"). These methods allow for precise control over facial expressions and attributes.
%
%

%
Recent progress in this field has led to more fine-grained applications, particularly in avatar generation~\cite{deng2022learning, deng2022gram, chan2022efficient, zhuang2022mofanerf, yin2022nerfinvertor, chen2023panic, zhou2023cips, xiang2023gram, galanakis20233dmm} and human pose generation ~\cite{liu2021neural, su2021nerf, zhang20223d, zhao2022humannerf, weng2022humannerf, jayasundara2023flexnerf, chen2023veri3d, hu2023sherf, mu2023actorsnerf, chen2023gm}. A significant achievement of these technologies is their ability to produce high-fidelity animations of real subjects using only a limited number of input images. This breakthrough not only conserves resources but also opens up exciting research prospects, especially in fields like video gaming, augmented and virtual reality (AR/VR), and human-computer interaction.
%

\subsubsection{Spatial Transformation Editing}
\begin{figure*}[!htbp]
    \centering
    \includegraphics[width=\linewidth]{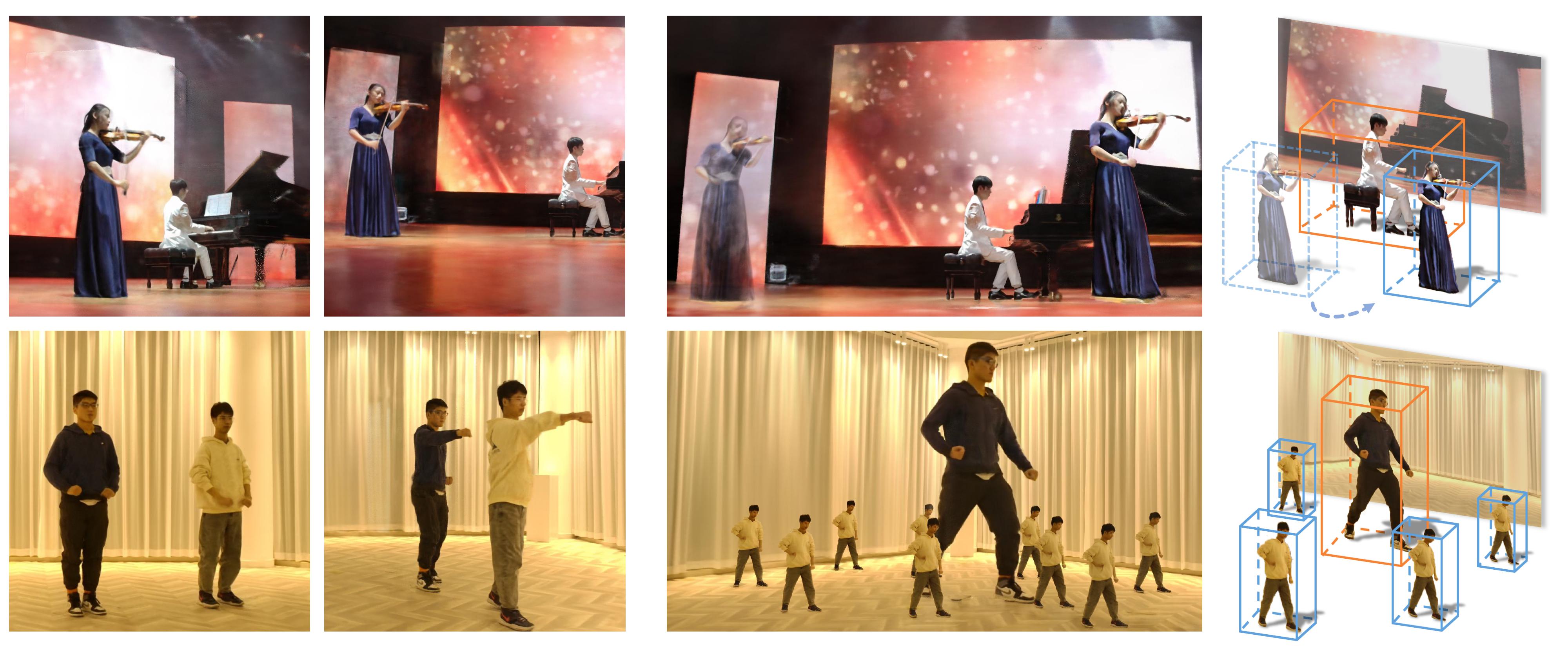}
\caption{Illustration of ST-NeRF's~\cite{zhang2021editable} capabilities to perform complex editing tasks in dynamic scenes. Here, we see the application of spatial affine transformations, temporal retiming, and transparency adjustments to selected objects within a scene. The transformations are applied across different timestamps and 3D bounding boxes around the target objects.}
%
    \label{fig:st_nerf_application}
\end{figure*}

\textbf{ST-NeRF}~\cite{zhang2021editable} presents a layered representation approach for each dynamic entity within scenes, where every entity is represented as a separate continuous function that spans both space and time. The MLP network of the model is composed of two key modules: a space-time deform module and a neural radiance module. In this setup, the frame number is directly encoded in the model. This approach to disentanglement of space and time facilitates various spatial editing techniques, such as affine transformation, insertion, and removal, along with temporal editing capabilities such as re-timing, as demonstrated in Figure~\ref{fig:st_nerf_application}.
%
%

Approaches like \textbf{AutoRF}~\cite{muller2022autorf}, \textbf{Neural Scene Graph}~\cite{ost2021neural}, \textbf{PNF}~\cite{kundu2022panoptic} or \textbf{DisCoScene}~\cite{xu2023discoscene} build a full 3D radiance field for each object contained within a bounding box. By treating objects' radiance fields as independent entities, we can render scenes more efficiently by focusing only on the relevant points where the rays intersect with these bounding boxes (ray-box intersection). This allows for image editing through the manipulation of bounding boxes, enabling the repositioning (rotation and translation) of objects in a scene without altering their visual appearance. For operations like removal or replication, users can adjust the scene's layout by deleting or cloning bounding boxes. In scenarios without bounding boxes~\cite{siddiqui2023panoptic}, object removal is accomplished by reducing the density of points associated with the target instance to zero. Meanwhile, duplicating the weights of an object's MLPs or its latent codes can result in the cloning of that instance in the scene.
%

%
\textbf{Control-NeRF}~\cite{lazova2023control} learns volumetric representations for multiple scenes by using a single shared rendering model. During testing, because the feature volumes are separate from the rendering model, the authors can perform spatial adjustments to these volumes or combine them. This process allows for the editing of the scene content without altering the fixed parameters of the rendering network.

\subsection{Object Detection and 6D Pose Estimation}
\label{sec:obj_det}
The task of 3D object detection is essential for a variety of applications, as it provides a detailed understanding of objects' sizes and positions in three dimensions. This task is more complex than 2D object detection due to the challenges in obtaining precise 3D data and the additional degrees of freedom (DoF). Methods based on point cloud representations depend heavily on accurate data from specialized sensors. Therefore, innovative techniques are necessary to leverage the capabilities of NeRFs while addressing the complexities of accurate 3D object detection from 2D images.
%

\subsubsection{3D Object Detection}
\textbf{NeRF-RPN}~\cite{hu2023nerf} is designed to identify all bounding boxes in a scene. The process begins by sampling a grid of points, from which RGB and density values are extracted using a pre-trained NeRF model. These volumetric features are then processed through a 3D Feature Pyramid Network (FPN)~\cite{lin2017feature} backbone, yielding deep, multi-scale 3D features. These features are inputted into a 3D Region Proposal Network (RPN) head, generating region proposals. A key innovation in NeRF-RPN is its use of a novel voxel representation, integrating multi-scale 3D neural volumetric features. This allows for the direct regression of 3D bounding boxes within NeRF without needing to render from any viewpoint.
In contrast, \textbf{NeRF-Det}~\cite{xu2023nerf}, a joint NeRF-and-Det method, links the NeRF branch with the detection branch using a shared geometry-based MLP. This setup enables the detection branch to use the gradient flow from NeRF in estimating the opacity fields. Consequently, it effectively masks out free space and reduces ambiguity in the feature volume, offering improvements over the NeRF-to-Det approach.
%

%
\textbf{MonoNeRD}~\cite{xu2023mononerd} approaches the concept of monocular 3D detection with NeRFs by considering intermediate frustum representations as SDF-based (Signed Distance Function-based) NeRFs. These are then optimized using volume rendering techniques. The process involves grid sampling on these frustum features to construct regular 3D voxel features along with corresponding densities. These voxel features are subsequently inputted into detection modules. This methodology establishes a new standard in monocular 3D detection using NeRFs.

On another front, techniques like \textbf{Neural groundplans}~\cite{sharma2022neural} and \textbf{SUDS}~\cite{turki2023suds} use feature field clustering to derive object-centric 3D representations in an unsupervised manner. These methods start with a dynamic field and apply traditional connected-component labeling in the feature space, considering the cumulative density values. This process aids in identifying individual objects. The smallest box enclosing each connected component is then computed, resulting in a 3D bounding box for every detected object.
%
%
%

\subsubsection{6D Pose Estimation}
\textbf{ShAPO}~\cite{irshad2022shapo} extracts comprehensive 3D details of multiple objects from a single RGB-D observation. This includes the objects' shape, 6D pose, scale, and appearance. The technique uses an octree-based differentiable optimization, drawing on pose, texture, and masks derived from an FPN~\cite{lin2017feature} backbone.
\textbf{NCF}~\cite{huang2022neural} is a method that estimates the 6D pose of a rigid object using a single RGB image. It maps from the camera space to the object model space. NCF predicts the corresponding 3D point in the model space and its signed distance. This facilitates the creation of 3D-3D correspondences, crucial for determining the object's pose.
\textbf{NeurOCS}~\cite{min2023neurocs} focuses on predicting the object mask and NOCS (Normalized Object Coordinate Space) map, which are then used in PnP (Perspective-n-Point) algorithms to estimate object pose. Additionally, a separate detector is applied to the NOCS and predicted depth data, aiding in precise 3D object localization.

\begin{figure}[!htbp]
    \centering
    \includegraphics[width=\linewidth]{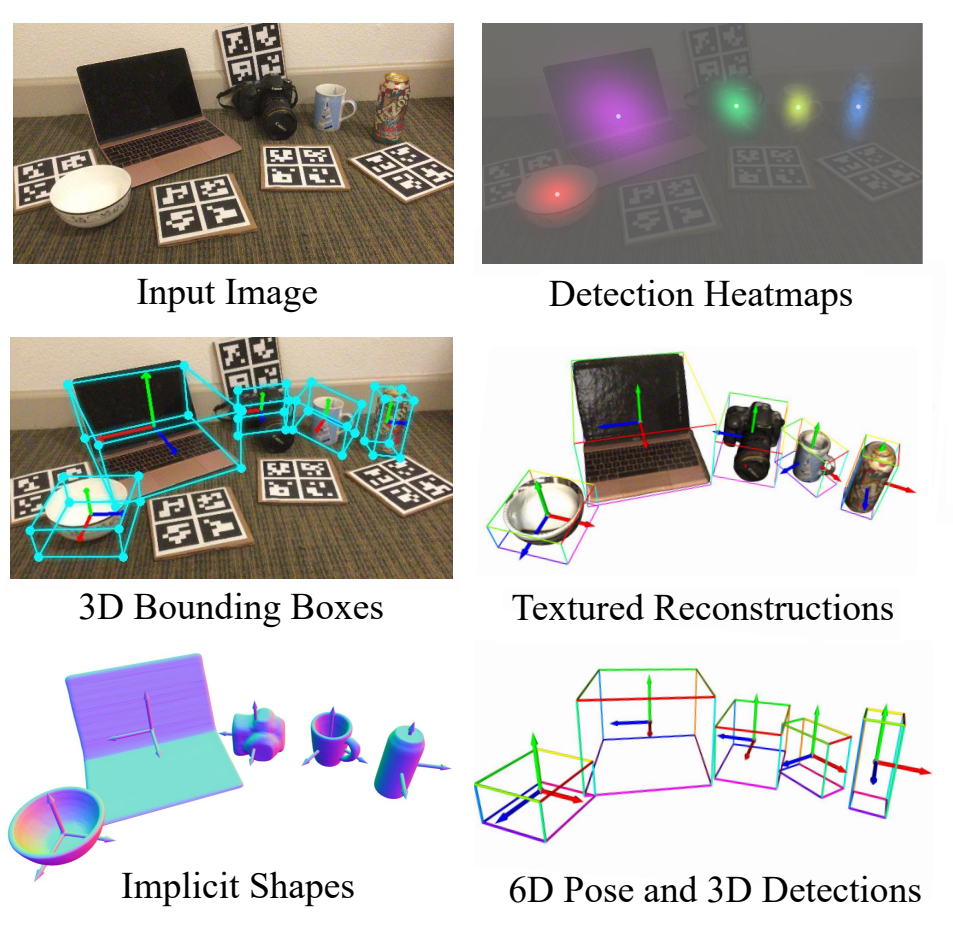}
\caption{ShAPO's~\cite{irshad2022shapo} multifaceted 3D object detection and pose estimation capabilities illustrated through an input image's transformation into detection heatmaps, 3D bounding boxes, textured reconstructions, and implicit shape representations.}
    \label{fig:shapo_application}
\end{figure} 
\begin{figure*}[!htbp]
    \centering
    \includegraphics[width=\linewidth]{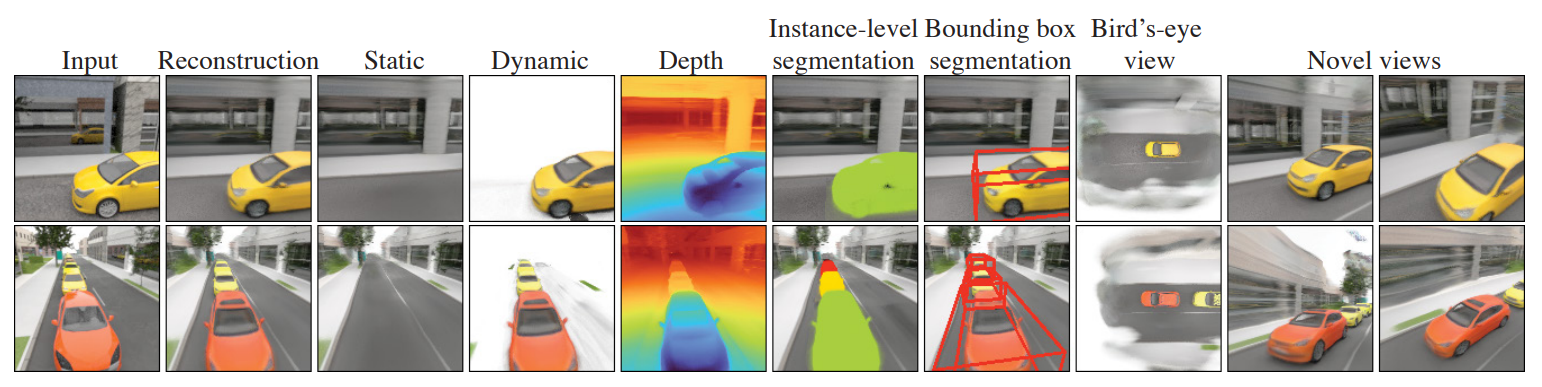}
\caption{\textbf{Semantically-trained Neural Groundplans~\cite{sharma2022neural} for Urban Scene Analysis} -- The model discerns static/dynamic entities and object instances, while enhancing novel view synthesis. Key features include synthesis of novel viewpoints, filling in missing information, accurate 3D bounding box predictions, 3D scene editing capabilities, and extraction of object-centric 3D representations.}   
%
    \label{fig:groundplans_application}
\end{figure*}
\subsection{Holistic Decomposition}\label{sec:decomposition}
\subsubsection{Objects vs Background}
\textbf{NeRF-W}~\cite{martin2021nerf} incorporates per-frame embeddings and a transient branch to model non-photometric consistent effects in unconstrained photo collections. Although it wasn't specifically designed to distinctly separate objects from their surroundings, it offers an innovative method for foreground element capture within various environments.

Subsequent research, including works~\cite{stelzner2021decomposing, xie2021fig, yang2021learning, bing2022dm}, has led to the division of NeRF into a dual-pathway architecture. This structure comprises a scene (background) branch for encoding scene geometry and appearance, and an object (foreground) branch for individual object encoding. These models learn to encode multiple objects simultaneously by assigning activation codes to training rays for each object, eliminating the need for separate training per object. For view generation with object manipulation, they render the transformed objects using the conditioned object branch and the background from the scene branch together. An added feature includes an object manipulator for precise radiance and object field editing, taking into account challenges such as object collisions and occlusions. Meanwhile, works like \textbf{uORF}'s~\cite{yu2022unsupervised, smith2023unsupervised}, aim to deduce latent object-centric representations into distinct slots through an attention mechanism, facilitating unsupervised segmentation.
%
%

In their panoramic room capture study, Yang \etal~\cite{yang2022neural} initially predict object metadata and infer object-to-object and object-to-room relationships, leveraging object-level predictions and geometric cues. They also incorporate pre-convolved HDR maps and surface normals into a global optimization, enabling the synthesis of novel lighting conditions and scene touring.
Zhu\etal~\cite{zhu2023occlusion} employ one MLP to accurately model a scene with occlusions and another MLP for the background. They train the background MLP and remove occlusions from aggregated information from scene MLP to determine if the output of the background NeRF matches the observed color, learning a mask from the ray's weights. This approach includes a depth constraint for probing occluded areas, by comparing the depth of occlusion and background, based on the assumption that occlusions are foreground objects at a closer distance.

\textbf{vMAP}~\cite{kong2023vmap} designs a vectorized object-level mapping, where each object is detected through its 3D point cloud and instance segmentation map, which is then represented by a separate MLP. The 3D bounds are continually updated, via data association across frames, leading to improvements in object-level reconstruction quality and runtime efficiency compared to traditional SLAM systems.

Zhang \etal~\cite{zhang2023nerflets} focus on representing scenes using small local radiance fields, termed ``nerflets''. Each nerflet covers a specific scene portion, determined by its influence function. These nerflets can collectively represent complex object instances, providing a more efficient and compact representation for outdoor environments that can be rendered, decomposed, and edited.

\textbf{AssetField}~\cite{xiangli2023assetfield} presents a natural visualization of scenes in Bird's Eye View (BEV) using informative ground feature planes aligned with the physical ground. This approach extracts and categorizes neural representations of scene objects, enabling users to manipulate and compose assets directly on the ground feature plane using feature patches from multiple scenes.

Haughton \etal~\cite{haughton2023real} and Chen \etal~\cite{chen2023perceiving} demonstrate how robots can identify objects and build composed 3D representations through physical interactions like pushing, grasping, or poking. The coherence of their model allows for the efficient propagation of measured physical properties (e.g., poses, rigidity, material) throughout the scene. Their experiments highlight the potential for automated sorting and grasping tasks.
%

\begin{figure}[!htbp]
     \centering
     \begin{subfigure}[b]{\linewidth}
     \begin{tikzpicture}[
     line/.style = {draw=black, line width=0.1mm},
     method/.style = {text width=3cm}]
     \draw[line] (-1.5,0) node[left, text=red]{\textbf{Overfitting}} 
                         --(1.5,0) node[right, text=red]{\textbf{Generalizable}};
     \draw[line] (0,-2.5) node[below, text=red]{\textbf{Scene-centric}}
                         --(0,2.5) node[above, text=red]{\textbf{Object-centric}};
     \node[method, align=right] at (2,2.5) {\hfill\cite{irshad2022shapo}};
     \node[method, align=right] at (2,2) {\hfill\cite{jang2021codenerf}\cite{liu2021editing}};
     \node[method, align=right] at (2,1.5) {\hfill\cite{rematas2021sharf}\cite{muller2022autorf}\cite{kim2022ae}};
     \node[method, align=right] at (2,1) {\hfill\cite{xie2021fig}};
     \node[method, align=right] at (2,0.5) {\hfill\cite{sharma2022neural}\cite{irshad2023neo}};

     \node[method, align=left] at (-2,2.5) {\cite{niemeyer2020differentiable}};
     \node[method, align=left] at (-2,2) {\cite{guo2020object}};
     \node[method, align=left] at (-2,1.5) {\cite{zhang2021editable}};
     \node[method, align=left] at (-2,1) {\cite{ost2021neural}\cite{kong2023vmap}\cite{xiangli2023assetfield}};
   
     \node[method, align=left] at (-2,0.5) {\cite{yang2021learning}\cite{bing2022dm}\cite{wang2023learning}};
     \node[method, align=left] at (-2,-0.5) {\cite{kundu2022panoptic}\cite{krishnan2023lane}\cite{wu2023mars}};
     \node[method, align=left] at (-2,-1.5) {\cite{siddiqui2023panoptic}\cite{fu2022panoptic}};
     \node[method, align=left] at (-2,-2) {\cite{martin2021nerf}};
     \node[method, align=left] at (-2,-2.5) {\cite{mildenhall2020nerf}\cite{barron2022mip}};

     \node[method, align=right] at (2,-0.5) {\hfill\cite{yu2022unsupervised}\cite{xu2023discoscene}};
     \node[method, align=right] at (2,-1) {\hfill\cite{lin2023componerf}\cite{cohen2023set}};
     \node[method, align=right] at (2,-2) {\hfill\cite{zarzar2022segnerf}\cite{liu2023semantic}};
     \node[method, align=right] at (2,-2.5) {\hfill\cite{yu2021pixelnerf}\cite{chibane2021stereo}\cite{lin2023vision}};
     \end{tikzpicture}
     
     \end{subfigure}
        
     \begin{subfigure}[b]{\linewidth}
         \centering
         \resizebox{\textwidth}{!}{
         \begin{tabular}{
         |lll|}
         \hline
        DVR~\cite{niemeyer2020differentiable} & Gou \etal~\cite{guo2020object} & STNeRF~\cite{zhang2021editable} \\
        NSG~\cite{ost2021neural} & vMap~\cite{kong2023vmap} & AssetField~\cite{xiangli2023assetfield} \\
        Object-NeRF~\cite{yang2021learning} & DM-NeRF~\cite{bing2022dm} & UDC-NeRF~\cite{wang2023learning} \\
        PNF~\cite{kundu2022panoptic} & LANe~\cite{krishnan2023lane} & MaRS~\cite{wu2023mars} \\
        Panoptic Lifting~\cite{siddiqui2023panoptic} & Panoptic NeRF~\cite{fu2022panoptic} & NeRF-W~\cite{martin2021nerf} \\
        NeRF~\cite{mildenhall2020nerf} & MipNeRF360~\cite{barron2022mip} & ShAPO~\cite{irshad2022shapo}\\
        CodeNeRF~\cite{jang2021codenerf} & EditNeRF~\cite{liu2021editing} & SHaRF~\cite{rematas2021sharf}\\
        AutoRF~\cite{muller2022autorf} & AE-NeRF~\cite{kim2022ae} & FigNeRF~\cite{xie2021fig}\\
        NGP~\cite{sharma2022neural} & Neo-360~\cite{irshad2023neo} & uORF\cite{yu2022unsupervised}\\
        DisCoScene~\cite{xu2023discoscene} & CompoNeRF~\cite{lin2023componerf} & Set-the-Scene\cite{cohen2023set}\\
        SegNeRF~\cite{zarzar2022segnerf} & PixelNeRF~\cite{yu2021pixelnerf} & S-RF~\cite{chibane2021stereo} \\
        VisionNeRF~\cite{lin2023vision} & &\\
         \hline
       \end{tabular}}
     \end{subfigure}
     \caption{Categorization of methods that jointly infer appearance and shapes.}
     \label{fig:taxonomy2}
\end{figure}
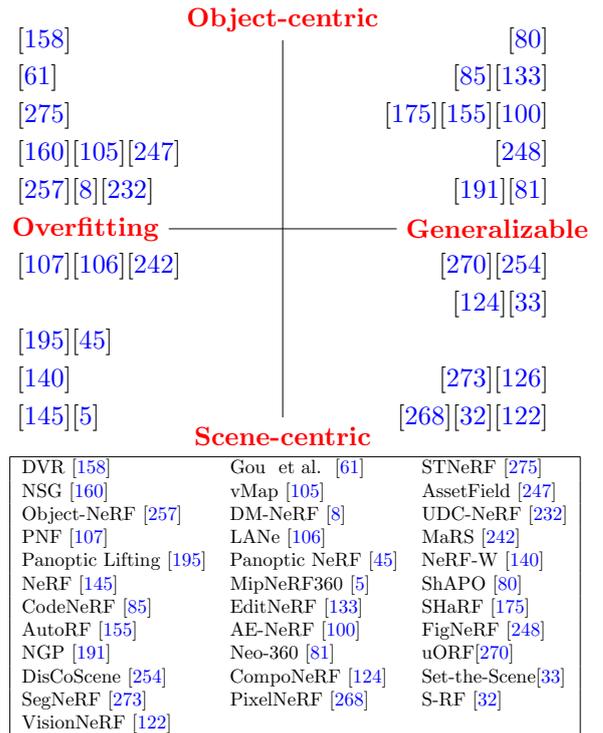
    
\subsubsection{Static vs Dynamic Objects}
Perceiving and representing dynamic environments is essential for autonomous agents to understand and interact with their surroundings. The key challenge involves disentangling camera and object motion while simultaneously reconstructing the dynamic scene. Such representations permit the synthesis of novel views within dynamic settings or the distinction between moving and stationary elements, offering flexibility in perspective and timing, i.e., in a free-view and time-varying manner.
Incorporating a temporal element into an MLP could serve as a viable approach. This would involve encoding the time variable $t$ , either by mapping it to a higher-dimensional space using frequency encoding or a 4D-hash grid, in a similar manner to the spatial coordinates $\mathbf{x}$ and $\mathbf{d}$, or through learnable, time-dependent latent codes as suggested in several studies~\cite{li2022neural, park2023temporal}.

For scenes that are predominantly static, optimizing a single model could lead to blurry outputs and inconsistencies. Solutions like \textbf{DynNeRF}~\cite{gao2021dynamic} and \textbf{STaR}~\cite{yuan2021star} have been developed to segregate moving objects from the static background. They use two separate branches for static landscapes and dynamic objects: a static branch containing non-moving topography consistent across videos and a dynamic branch that handles dynamic objects. The training of these branches is often directed using pre-existing semantic and motion segmentation methods, creating masks that exclude ``dynamic'' pixels from the static training process. This approach ensures the background is reconstructed accurately without conflicting the losses, avoiding errors caused by moving objects.
Additionally, temporal variations can be accounted for in a self-supervised manner through regularization~\cite{yuan2021star, tschernezki2021neuraldiff, sharma2022neural, wu2022d2nerf}, which enables the dynamic field to learn as necessary. \textbf{D$^{2}$NeRF}~\cite{wu2022d2nerf}, an extension of \textbf{HyperNeRF}~\cite{park2021hypernerf} to dynamic scenes, can handle complex scenes involving multiple non-rigid and topologically varying objects. This method is able to decouple dynamically moving shadows with a separated field that reduces the static radiance output as well. 
The features encoded from both branches can be regularized during training and can be interpolated using MLPs or 4D hash grids in both short- and long-term space-time ranges~\cite{park2023temporal}. This technique not only delivers high-quality, mooth rendering performance but also enhances the efficiency and stability of the training process.

On the contrary, the dynamic model proposed by Li in \textbf{NSFF}~\cite{li2021neural} goes a step further by directly predicting forward/backward scene flows along with disocclusion weights from a multilayer perceptron (MLP). These disocclusion weights act as an unsupervised confidence, determining the locations and intensity at which to apply the temporal photoconsistency loss. The model uses a pre-trained 2D optical flow model to supervise the predicted 3D flows, which are also refined using a cycle consistency term for regularization. Building on this work, \textbf{SAFF}~\cite{liang2023semantic} enhances the model by also generating semantic and saliency features, which are instrumental in refining the segmentation of static and dynamic elements within the scene. 
In a similar vein, \textbf{Factored-NeRF}~\cite{wong2023factored} leverages annotations from keyframes, propagating them to adjacent frames to deduce scene flows, map object trajectories, and determine rigidities. Through comprehensive end-to-end optimization, this model gains the ability to modify object placements, trajectories, and even adapt to non-rigid movements. By computing static and dynamic fields independently, these methods facilitate the separate rendering of stationary and moving parts within a scene.
%
%
%
In addition, a significant body of research is dedicated to testing these approaches in dynamic environments, notably those with complex movements, such as vehicles and pedestrians in urban settings. \textbf{Neural Scene Graphs}~\cite{ost2021neural} introduces a learned scene graph representation that encodes the transformations and radiance of objects. This method uses tracking data and video frames to learn distinct representations for each object within the scene graph, thereby streamlining the process of synthesizing and decomposing views across various object arrangements and dynamic conditions. This progress enables not only the realistic rendering of new scenes and objects, but also the potential for 3D object detection through the technique of inverse rendering.
%

\textbf{PNF}~\cite{kundu2022panoptic}, \textbf{LANe}~\cite{krishnan2023lane}, and \textbf{MARS}~\cite{wu2023mars} also break down scenes into distinct objects and backgrounds, employing panoptic segmentation and bounding boxes for each object. Each object is represented by an oriented 3D bounding box and is characterized by a dedicated MLP that computes density and radiance from inputs like position, direction, and pose. These MLPs are tailored to individual instances and refined through a meta-learning initialization process~\cite{kundu2022panoptic}. 
\textbf{LANe}~\cite{krishnan2023lane} trains on a single scene under varying lighting conditions, learning to adapt by creating a light field and using a corresponding shader to modulate the appearances of objects for coherent integration into scenes with different lighting. 
\textbf{SUDS}~\cite{turki2023suds} and the subsequent \textbf{EmerNeRF}~\cite{yang2023emernerf} handle scalability by employing a multi-resolution hash table for scene partitioning, enabling dynamic management of vast numbers of objects over extensive areas (hundreds of kilometers), using implicit scene flows and DINO~\cite{caron2021emerging} features for enhancement. \textbf{Neural groundplans}~\cite{sharma2022neural} process their ground-aligned 2D feature grids through a 2D CNN, effectively disentangling the representation into two distinct groundplans for static and dynamic features, thus achieving a clear disentanglement.
%
%

\subsection{NeRFs and Language}
\label{sec:language}
\subsubsection{Text-driven 3D Generation and Editing}
Text-guided image generation has seen tremendous success in recent years, primarily due to the breathtaking progress
in language image and diffusion models. These have also inspired major breakthroughs in text-guided shape generation.
This progress has influenced research, linking NeRFs with textual input descriptions.

\textbf{CLIP-NeRF}~\cite{wang2022clip} extends the work on conditional Neural Radiance Fields (NeRF) by promoting similarity between the CLIP~\cite{radford2021learning} embeddings of scenes, facilitating user-friendly manipulation of NeRF through short text prompts or example images. This approach disentangles the latent representation, allowing for separate control over the shape and appearance of objects. Consequently, it enables the creation of code mappers that modify latent codes based on user-specified edits via text prompts or images, demonstrating improved editing capabilities and narrowing the gap between textual and visual editing cues.
%
%
\textbf{DreamField}~\cite{jain2022zero} employs a CLIP model pre-trained on large datasets of captioned images from the web. It guides the generation process so that the rendered images achieve high scores with a target caption according to the CLIP model, even without access to 3D shape or multi-view data. This method facilitates the zero-shot generation of diverse 3D objects from captions. Additionally, Lee \etal~\cite{lee2022understanding} explored the performance of different CLIP model architectures in voxel grid representations, finding that an ensemble of models for guidance can prevent adversarial generations and improve geometrical structure, memory, and training speed.
%

%
In parallel, the application of 2D diffusion models for similar purposes, as discussed in~\cite{rombach2022high}, is introduced. Since NeRFs operate in image space, guiding a NeRF scene with the diffusion model involves practical solutions like deriving a Score Distillation loss or leveraging the training process in latent space, as seen in \textbf{DreamFusion}~\cite{poole2022dreamfusion} and \textbf{Latent-NeRF}~\cite{metzer2023latent}. However, these approaches often result in unsatisfactory outputs and low diversity in objects generated from the same input text, coupled with lengthy synthesis times.
%
%
Addressing these challenges, \textbf{DITTO-NeRF}~\cite{seo2023ditto} introduces progressive reconstruction schemes focusing on scales (from low to high resolution), angles (from inner to outer boundaries initially), and masks (from object to background boundary). This methodology achieves significant improvements in diversity and quality, as well as speed and fidelity of the generated objects, marking a significant progress in the field.
%

%
\textbf{LaTeRF}~\cite{mirzaei2022laterf} enhances the NeRF framework by incorporating an ``objectness'' probability for each point, allowing the extraction of objects from scenes using pixel annotations. 
\textbf{SINE}~\cite{bao2023sine} enhances semantic editing by introducing advanced methods: cyclic constraints alongside a proxy mesh for accurate geometric modifications, a color compositing system for better texture editing, and feature cluster-based regularization to manage the edited areas while maintaining the integrity of content that is not being edited.
These enhancements facilitate compatibility with off-the-shelf text-prompt editing methods, enabling modifications to an object's appearance and geometry, and the inpainting of missing parts of an object based on textual cues.
%
%
%
\textbf{NeRF-Art}~\cite{wang2023nerf} and \textbf{Blending-NeRF}~\cite{song2023blending} integrate a pre-trained NeRF with an editable NeRF. The editable NeRF is trained to render a blended image that aligns with a target text, allowing precise editing of 3D object regions while preserving their original appearance.
%
%
\textbf{Instruct-NeRF2NeRF}~\cite{haque2023instruct} iteratively updates dataset images during NeRF model training with global text instructions from InstructPix2Pix~\cite{brooks2023instructpix2pix}. This process, involving a loss that combines rays from various viewpoints, leads to higher quality results and more stable optimization.

%
Existing methods, however, face limitations in controlling individual objects within a scene. Modifying specific scene aspects without affecting others remains a challenge, and scene-level editing with long text prompts can lead to guidance collapse, preventing specific scene component edits. \textbf{CompoNeRF}~\cite{lin2023componerf} and \textbf{Set-the-Scene}~\cite{cohen2023set} address these issues by employing a composition module to adjust text guidance levels and ensure the distinctiveness of entities while maintaining overall scene coherence. They represent the scene as a composition of multiple NeRFs, each optimized to represent specific objects ``locally'' and integrate seamlessly into the broader scene ``globally'', thus eliminating guidance ambiguity. Through proxy manipulation, scenes can be decomposed and reassembled for editing without the need for additional fine-tuning.
%

\subsubsection{Queryable Interaction}
\textbf{CLIP-Fields}~\cite{shafiullah2022clipfields} integrates the strengths of the CLIP~\cite{radford2021learning} image encoder, Sentence BERT~\cite{reimers2019sentence}, and NeRF to create a 3D scene representation that is queryable for mobile robots. 
This architecture is equipped with heads that output vectors corresponding to natural language descriptions, the visual appearance of objects, and the instance identification of every specific point in space. 
It uses two contrastive losses: one for the label token and another for the visual language embedding. 
CLIP-Fields demonstrate robustness in low-shot scenarios and label errors, capable of answering queries with varying degrees of real-life complexity.
\textbf{VL-Fields}~\cite{tsagkas2023vlfields} aims to overcome the limitations of CLIP-Fields, which are restricted to a subset of scene points with known object classes. It proposes an open-set visual-language model that operates without prior knowledge of the object classes present in the scene.
%

%
\textbf{LERF}~\cite{kerr2023lerf} uses a multi-scale feature pyramid combining 3D CLIP field and DINO~\cite{caron2021emerging} features to refine object boundaries for language query interactions. It allows for pixel-aligned queries of distilled 3D CLIP embeddings, bypassing the need for region proposals, masks, or fine-tuning. LERF supports hierarchical, long-tail, open-vocabulary queries across the scene volume.
\textbf{F3RM}~\cite{shen2023F3RM} conducts few-shot learning experiments for grasping and placing tasks, drawing on Deep Fusion Field~\cite{kobayashi2022decomposing} (DFF) methodologies. This enables robots to perform 6-DoF object manipulation in response to natural language commands, exhibiting open-set generalization capabilities for handling unseen objects with significant differences.
\textbf{GNFactor}~\cite{ze2023gnfactor} optimizes a generalizable NeRF for reconstruction alongside a Perceiver Transformer~\cite{jaegle2021perceiver} for decision-making. This transformer integrates the robot's proprioception and language features to execute decisions based on a Q-function~\cite{mnih2013playing}, facilitating advanced decision-making processes in robotic applications.
%

\section{Datasets and Evaluation}
\label{sec:evaluation}

\subsection{Core Metrics and Principles}
\subsubsection{Reconstruction and Novel View Synthesis}
Image reconstruction and novel view synthesis in the standard setting use visual quality assessment metrics for benchmarks. The following metrics are the common standards in the NeRF literature:

\textbf{Peak Signal-to-Noise Ratio (PSNR)} quantifies the ratio of the maximum possible signal power, represented by the highest pixel intensity value, to the power of the noise corrupting the signal. A higher PSNR value indicates superior image quality. However, PSNR may not reliably reflect perceptual similarity since it fails to precisely represent how humans perceive image quality.

%
The \textbf{Structural Similarity Index Metric (SSIM)}~\cite{wang2004image} offers a perceptually more relevant evaluation by comparing two images through aspects such as luminance, contrast, and structural integrity. It considers variations in pixel intensities, spatial relationships, and texture contrasts. SSIM values span from -1 to 1, with 1 signifying an exact correspondence between the original and the reconstructed images. In terms of aligning with human visual perception, SSIM delivers a more accurate measure of image quality than PSNR.

The \textbf{Learned Perceptual Image Patch Similarity (LPIPS)}~\cite{zhang2018unreasonable} metric assesses perceptual similarity between rendered views/poses and their corresponding ground truth images from specific viewing directions. Using deep learning, this perceptual metric measures the similarity between two images based on features extracted from a pre-trained Convolutional Neural Network (CNN), such as AlexNet or VGG, trained on the ImageNet dataset. Designed to more closely mirror human perception of image similarity, lower LPIPS scores indicate a greater perceptual similarity between the compared images. LPIPS proves to be especially effective in identifying subtle geometric and textural differences, making it particularly valuable for evaluating generative models and tasks related to image synthesis.

To enable easier comparison, an ``average'' error metric that summarizes all three above metrics is supplementarily presented~\cite{barron2021mip}:\\
$\text{Average}=\sqrt[3]{10^{-\text{PSNR}/10}.\sqrt{1-\text{SSIM}}.\text{LPIPS}}$

\textbf{Fréchet Inception Distance (FID)}~\cite{heusel2017gans} is a metric used to measure the similarity between the distribution of real images and the distribution of generated images in feature space. It uses the Inception-v3~\cite{szegedy2016rethinking} model to extract features from real and generated images. The FID score is calculated by computing the Fréchet distance between the multivariate Gaussian distributions of the feature representations of real and generated images. A lower FID score indicates that the generated images are more similar to the real images in terms of visual appearance and diversity.

\textbf{Kernel Inception Distance (KID)}~\cite{binkowski2018demystifying} is an extension of the FID that aims to address some limitations of the FID. It measures the Maximum Mean Discrepancy between the feature distributions of real and generated images using kernel functions. KID focuses on a more robust and informative evaluation of image quality and diversity by considering the distributional properties of the features. It provides an unbiased estimation of the true distance between distributions of real and generated images, ensuring a more accurate representation of their similarity in feature space. Moreover, KID's robustness to the choice of sample size minimizes the variability stemming from different sample sizes and requires fewer samples for calculation compared to alternative metrics.

\subsubsection{Segmentation}
Various evaluation metrics are employed to assess the performance of segmentation algorithms, quantifying the accuracy and reliability of the delineation between different regions in an image. Here are some of the common metrics used:

\textbf{Pixel Accuracy} computes the proportion of correctly classified pixels over the total number of pixels. It's a simple and intuitive measure but might not capture the overall performance accurately, especially when dealing with imbalanced classes.

\textbf{Mean Intersection over Union (mIoU)}, also referred to as the Jaccard index, is essentially a method to quantify the percent overlap between the target mask and our prediction output. mIoU is calculated by taking the IoU of each class and averaging them.
\begin{equation}
\text{IoU}(p,g) = \frac{|p\cap g|}{|p \cup g|}
\label{eq:iou}
\end{equation}

Panoptic segmentation combines both semantic segmentation and instance segmentation. As a result, evaluation metrics for panoptic segmentation are crucial for quantitatively assessing the performance of algorithms that classify each pixel in an image into predefined classes or instance IDs and need to consider both aspects.
\textbf{Panoptic Quality (PQ)}~\cite{kirillov2019panoptic} is defined as the average IoU of the matched segments, while the denominator (see Equation \ref{eq:pq}) is designed to penalize segments without matches. PQ treats the quality of segmentation masks for all classes in an interpretable and unified manner, capturing all aspects of the task.
\begin{equation}
\text{PQ} = \frac{\sum_{(p,g)\in TP}\text{IoU}(p,g)}{|TP|+\frac{1}{2}|FP|+\frac{1}{2}|FN|}
\label{eq:pq}
\end{equation}

\subsection{Public Datasets for Semantically-aware NeRFs}

Existing datasets for novel view synthesis in the classical NeRF literature can be grouped into the following major categories:
\begin{itemize}
    \item [a] \underline{Hemispherical 360° inward-facing views} around an object of interest, which is mostly set against a plain white background (these include ShapeNet~\cite{chang2015shapenet}, CO3D~\cite{reizenstein2021common}, OmniObject3d~\cite{wu2023omniobject3d}, and Realistic Synthetic~\cite{mildenhall2020nerf}).

    \item [b] \underline{Forward-facing scenes} which aim the camera in a single direction and move in the vicinity facing the object (these include DTU~\cite{jensen2014large, aanaes2016large} and LLFF~\cite{mildenhall2019llff}).

    \item [c] \underline{Unbounded 360° real-scenes} that provide full surrounding coverage with detailed backgrounds (these include Tanks and Templates~\cite{knapitsch2017tanks} and MipNeRF360~\cite{barron2022mip} dataset).
\end{itemize}

Although fine-grained reconstruction is possible with provided camera intrinsics and poses, these datasets often lack compositional annotations (such as 3D bounding boxes or multi-object masks) and usually include a limited number of scenes. Efforts have been made to minimize photometric variation and avoid introducing multiple complex objects during capture. However, radiance field scene representations trained on these datasets typically focus on individual, per-scene optimization without additional semantic annotations or learning generalized priors. This makes it challenging to evaluate the performance of most semantically aware NeRFs.

Certain methods use hand-crafted annotations or pre-trained models to extract regions of interest from the scenes, but these approaches lack reliability as official benchmarks for comparing different methods. Therefore, in this section, we will discuss publicly available datasets that contain high-quality data with semantic annotations and which are the most relevant and most widely used in the literature.

\newcommand{\Dhref}[3][blue]{\href{#2}{\color{#1}{#3}}}%
\begin{table*}
    \centering
    \renewcommand{\arraystretch}{2}
    \centerline{\resizebox{1.1\textwidth}{!}{\begin{tabular}{|l|c|c|c|c|c|c|l|c|}
        \hline
        \textbf{Datasets} & \textbf{Venue} & \textbf{\#Scenes} & \textbf{\#Imgs} & \textbf{Centricity} & \textbf{Type} & \textbf{Data Modalities} & \textbf{Annotations} & \textbf{URL}\\
        \hline
        3DMV-VQA~\cite{hong20233d} & CVPR 2023 & 5000 &  600K & S+O & Indoor & RGB & Visual question \& answer & \Dhref{https://vis-www.cs.umass.edu/3d-clr/}{\faLink}\\
        \hline
        NeRDS 360~\cite{irshad2023neo} & ICCV 2023 & 75 & 15k & S+O & Urban & Synthetic & \makecell[l]{3D object boxes\\2D panoptic segmentation} & \Dhref{https://zubair-irshad.github.io/projects/neo360.html}{\faLink}\\
        \hline
        ScanNet++~\cite{yeshwanth2023scannet} & ICCV 2023 & 460 & 3.7M & S & Indoor & RGB-D & \makecell[l]{2D/3D panoptic segmentation} & \Dhref{https://cy94.github.io/scannetpp/}{\faLink}\\
        \hline        
        KITTI-360~\cite{liao2022kitti} & PAMI 2022 & 10 & 150K & S+O & Urban & RGB \& LiDAR & \makecell[l]{2D/3D object boxes\\2D panoptic segmentation} & \Dhref{https://www.cvlibs.net/datasets/kitti-360/}{\faLink}\\
        \hline
        SHIFT~\cite{sun2022shift} & CVPR 2022 & 4850 & 2.5M & S+O & Urban & Synthetic & \makecell[l]{2D/3D object boxes\\2D panoptic segmentation} & \Dhref{https://www.vis.xyz/shift/}{\faLink}\\
        \hline
        HM3D Sem~\cite{yadav2023habitat} & arXiv 2022 & 216 & - & S & Indoor & Mesh & 3D semantic segmentation & \Dhref{https://aihabitat.org/datasets/hm3d-semantics/}{\faLink}\\
        \hline
        3D-FRONT~\cite{fu20213d} & ICCV 2021 & 18968 & - & S+O & Indoor & Synthetic & 3D semantic segmentation & \Dhref{https://tianchi.aliyun.com/specials/promotion/alibaba-3d-scene-dataset}{\faLink}\\
        \hline
        HyperSim~\cite{roberts2021hypersim} & ICCV 2021 & 461 & 77.4K& S+O & Indoor & Synthetic & \makecell[l]{2D/3D object boxes\\2D/3D panoptic segmentation} & \Dhref{https://github.com/apple/ml-hypersim}{\faLink}\\
        \hline
        Waymo~\cite{sun2020scalability} & CVPR 2020 & 1150 & 1M & S+O & Urban & RGB \& LiDAR & \makecell[l]{2D/3D object boxes\\2D panoptic segmentation} & \Dhref{https://waymo.com/open/}{\faLink}\\
        \hline
        nuScenes~\cite{caesar2020nuscenes} & CVPR 2020 & 1000 & 1.4M & S+O & Urban & RGB \& LiDAR & \makecell[l]{3D object boxes\\2D semantic segmentation} & \Dhref{https://www.nuscenes.org/}{\faLink}\\
        \hline
        Replica~\cite{straub2019replica} & arXiv 2019 & 18 & - & S & Indoor & Mesh & 2D/3D panoptic segmentation & \Dhref{https://github.com/facebookresearch/Replica-Dataset}{\faLink}\\
        \hline
        Matterport 3D~\cite{chang2017matterport3d} & 3DV 2017 & 90 & 194.4K& S & Indoor & RGB-D & 2D/3D panoptic segmentation & \Dhref{https://niessner.github.io/Matterport//}{\faLink}\\
        \hline
        CLEVR~\cite{johnson2017clevr} & CVPR 2017 & - & 100K& O & Indoor & Synthetic & Visual question \& answer & \Dhref{https://cs.stanford.edu/people/jcjohns/clevr/}{\faLink}\\
        \hline
        ScanNet~\cite{dai2017scannet} & CVPR 2017 & 1513 &  2.5M & S+O & Indoor & RGB-D & \makecell[l]{3D object boxes\\2D/3D panoptic segmentation} & \Dhref{http://www.scan-net.org/}{\faLink}\\
        \hline
        Virtual KITTI~\cite{gaidon2016virtual} & CVPR 2016 & 5 & 17K& S+O & Urban & Synthetic & \makecell[l]{2D/3D object boxes\\2D panoptic segmentation} & \Dhref{https://europe.naverlabs.com/research/computer-vision/proxy-virtual-worlds-vkitti-2/}{\faLink}\\
        \hline
        SUN RGB-D~\cite{song2015sun} & CVPR 2015 & 47 & 10.3K& S+O & Indoor & RGB-D & \makecell[l]{2D/3D object boxes\\2D panoptic segmentation} & \Dhref{https://rgbd.cs.princeton.edu/}{\faLink}\\
        \hline        
        Shapenet~\cite{chang2015shapenet} & arXiv 2015 & - & - & O & Objects & CAD model & 3D part segmentation & \Dhref{https://shapenet.org/}{\faLink}\\
        \hline
        KITTI~\cite{geiger2012we, geiger2013vision} & CVPR 2012 & 22 & 15K& S+O & Urban & RGB \& LiDAR & \makecell[l]{2D/3D object boxes\\2D panoptic segmentation} & \Dhref{https://www.cvlibs.net/datasets/kitti/}{\faLink}\\
        \hline
    \end{tabular}}}
    \caption{\textbf{Overview of existing datasets for SRF-based multi-view scene understanding.} `Centricity' refers to scene and/or object-centric datasets, respectively denoted with S and O above.}
    \label{tab:my_label}
\end{table*}





\subsubsection{Indoor Scenes}
\textbf{Scannet}~\cite{dai2017scannet} is an RGB-D video dataset containing 2.5M views stemming from more than 1,500 scans, annotated with 3D camera poses, surface reconstructions, and instance-level semantic segmentations. It includes both 2D and 3D data and supports several 3D scene understanding tasks, including 3D object classification, semantic voxel labeling, and CAD model retrieval.

\textbf{Replica}~\cite{straub2019replica} consists of 18 highly photorealistic 3D indoor scene reconstructions at room and building scale. Each scene consists of a dense mesh, high-resolution HDR textures, per-primitive semantic class and instance information, and planar mirror and glass reflectors.

\textbf{Hypersim}~\cite{roberts2021hypersim} is a photorealistic synthetic dataset for holistic indoor scene understanding. It contains 77,400 images of 461 indoor scenes with detailed per-pixel labels and corresponding ground truth geometry, including complete scene geometry, material information, lighting information for every scene, dense per-pixel semantic instance segmentations, and complete camera information for every image.



\textbf{HM3DSem}~\cite{yadav2023habitat} consists of 142,6K annotations of object instances in 216 spaces and 3,100 rooms within those spaces, built on top of Matterport 3D~\cite{chang2017matterport3d} for Embodied AI applications. A key difference setting from other datasets is the use of texture information to annotate pixel-accurate object boundaries.

The most recently referenced dataset, \textbf{Scannet++}~\cite{yeshwanth2023scannet}, offers high-resolution and high-quality RGBD captures, supporting the novel view synthesis task along with dense semantic annotations. It encompasses 460 scenes, featuring 280K DSLR images and more than 3.7M iPhone RGBD frames.

\textbf{PeRFception}~\cite{jeong2022perfception} uses radiance fields (Plenoxels~\cite{yu2021plenoctrees, fridovich2022plenoxels}) as another representation of data that effectively conveys the same information for both 2D and 3D in a unified and compressed model, eliminating the need to store different data formats separately. At the moment, PeRFception-CO3D and PeRFception-ScanNet are created, which have covered object-centric and scene-centric environments respectively.

To tackle the problems of collecting, processing, and annotating datasets for 3D scene understanding at scale, \textbf{Kubric}~\cite{greff2022kubric} is introduced as a framework for generating synthetic datasets with fine-grain control over data complexity and rich ground truth annotations. The pipeline is linked with an open-source Python framework and Blender, allowing facilitating reuse of data-generation code, across multiple scales. Furthermore, Kubric can provide various randomization options for custom use cases. There have been many papers that applied this framework to create their own datasets~\cite{vora2022nesf, wu2022d2nerf, bhalgat2023contrastive, gao2023object}. However, most of the collected datasets are new in the field and limited within the approaches of the articles without proper benchmarks, they still remain as important parts and are waiting to be tested by the community.

\subsubsection{Outdoor Urban Scenes}
The \textbf{KITTI}~\cite{geiger2012we, geiger2013vision} dataset is a renowned collection tailored for computer vision research in the context of urban-scale 2D-3D environments, specifically designed to train and evaluate algorithms aimed at autonomous driving technologies. This dataset was compiled using raw LiDAR and video data collected in Karlsruhe, Germany, employing a vehicle-mounted system equipped with GPS and an inertial measurement unit. To accommodate various research objectives, parts of the dataset have been manually annotated by researchers, making KITTI a comprehensive resource that includes labeled data for a range of tasks such as stereo 2D-3D segmentation, optical flow, odometry, 2D-3D object detection, tracking, lane detection, and depth prediction/completion. However, the absence of complete semantic labeling limits its use for tasks like synthesizing new view images or constructing large-scale semantic maps, as these activities require fully labeled datasets for accurate evaluation.

Other datasets such as \textbf{nuScenes}~\cite{caesar2020nuscenes}, \textbf{Waymo}~\cite{sun2020scalability} try to address this shortcoming by providing more comprehensive data with semantic/instance labels in 2D and 3D, and richer 360° sensory information corresponding to longer driving logs with more accurate and geolocalized vehicle poses. Especially, \textbf{KITTI-360}~\cite{liao2022kitti} with its 3D-to-2D label transfer that opens more interesting tasks, \eg semantic SLAM or novel view semantic synthesis.

Adapting to a continuously evolving environment is a safety-critical challenge inevitably faced by all autonomous driving systems. Existing image and video-driving datasets, however, fall short of capturing the mutable nature of the real world. In other words, they are captured under approximately stationary conditions. \textbf{Virtual KITTI}~\cite{gaidon2016virtual} and \textbf{SHIFT}~\cite{sun2022shift} captures these driving scenarios in various environmental directions: time of day, cloudiness, rain, fog strength and vehicle and pedestrian density with more detailed object class annotations (persons, cars, license plates, ...) in separate discrete variations~\cite{gaidon2016virtual} or continuously shifting conditions~\cite{sun2022shift}.

\textbf{NERDS 360}~\cite{irshad2023neo} is a large-scale dataset for 3D urban scene understanding. This dataset consists of 75 outdoor urban scenes with diverse backgrounds in 360° hemispherical views, encompassing over 15K images. The dataset and the corresponding tasks are extremely challenging due to occlusions, diversity of backgrounds, and rendered objects with various lightning and shadows.

\subsubsection{Vision and Language}
\textbf{CLEVR}~\cite{johnson2017clevr} is a diagnostic dataset for studying the ability of Visual Question Answering (VQA) systems. It contains 100K rendered images and 853K generated unique questions for visual reasoning abilities such as counting, comparing, logical reasoning, and storing information. Each object present in the scene, aside from position, is characterized by a set of four attributes: 2 sizes: large, and small, 3 shapes: square, cylinder, and sphere, 2 material types: rubber, and metal, 8 color types: gray, blue, brown, yellow, red, green, purple, and cyan, resulting in 96 unique combinations.

\textbf{3DMV-VQA}~\cite{hong20233d} consists of approximately 5K scenes, and 600K images, paired with 50K questions in total. This dataset is built on top of the HM3DSem dataset~\cite{yadav2023habitat} with four types of questions: conceptual, counting, relational, and comparison questions. The authors further propose a 3D concept learning and reasoning framework that is grounded on open-vocabulary semantic concepts on 3D representation.

\section{Challenges and Perspectives}
\label{sec:challenges}

To progress in the field of semantically-aware NeRFs, targeted research efforts are essential. This section outlines the primary challenges and opportunities for enhancement that we have identified as critical focus areas.

\begin{itemize}
    \item [i] \textbf{Scene Generalizability.} 
Current Semantically-aware NeRF (SRF) methods, capable of processing datasets without the need for scene-specific training or optimization, mark a significant progress over the original NeRF methodology~\cite{mildenhall2020nerf}, which lacked any capability for cross-dataset generalization. Despite these improvements, there are still clear limitations. Current approaches might necessitate expensive, dense semantic annotations~\cite{irshad2023neo}, require substantial data volumes~\cite{sajjadi2022scene}, predominantly operate within synthetic settings, or produce blurriness in novel view synthesis, often attributed to L2 loss training~\cite{sajjadi2022scene,yu2021pixelnerf}. These challenges are further amplified by the variation in viewpoint densities during test-time, affecting traditional performance and efficiency.
Additionally, while some strategies employ pre-trained, sparse detectors and segmentation networks, they tend to achieve only object-centric generalization~\cite{muller2022autorf,guo2020object}. Addressing these challenges to improve cross-dataset generalization, or combining their respective benefits, would represent a substantial leap forward, enabling true real-time applications from acquisition to rendering.

    \item [ii] \textbf{Camera Calibration.} 
    While some NeRF-based methods are designed to take unposed images and simultaneously recover their extrinsic matrices~\cite{jeong2021self,wang2021nerf}, most NeRF derivatives work under the assumption that the RGB views provided as input are already posed. As a result, even minor calibration errors can lead to significant semantic misalignments across different views. Such misalignments can cause early failures in the process, which are often irreversible during the subsequent training or scene optimization phases. Therefore, there is a clear necessity within the NeRF domain as a whole to not only improve calibration techniques but also to develop specific mechanisms for pose refinement during the training process.
    
    \item [iii] \textbf{Data Efficiency and Augmentation.} Addressing the data efficiency challenge is essential for making NeRF more practical in real-world settings. Future work may involve exploring methods to train accurate semantically-aware models with less training data, and fewer annotations, making them more accessible for a broader range of applications, in particular in one/few-shot settings in real-world environments.
    The successful integration of semantic understanding into NeRFs has the potential to dramatically enhance applications in augmented reality, autonomous navigation, and beyond, by providing more meaningful and context-aware interpretations of 3D environments. These functionalities make NeRF well-suited to serve as the foundation for various components. By understanding the decomposition of the scene and allowing dynamic adjustments to the simulation environment, NeRF becomes a valuable asset in creating realistic scenarios for closed-loop simulations, providing a crucial element in training and testing scenarios for many systems. Additionally, its adaptability for data augmentation enhances its utility in improving the robustness and generalization of machine learning models.
    
    \item [iv] \textbf{Multi-modal, Multi-task, and Efficient Scene Understanding.} 
    Currently, the majority of multi-modal approaches investigated within the NeRF domain are centered around textTo3D. Despite the wide range of potential multi-task combinations available~\cite{zamir2018taskonomy,atanov2022task}, many remain largely unexplored, representing missed  opportunities to discover new, mutually informative tasks specifically within the area of Radiance Fields. For instance, areas such as sound processing or other types of inputs~\cite{luo2022learning} have yet to be fully explored.
    
    \item [v] \textbf{Real-Time and Mobile Performance.} NeRF encounters challenges in terms of computational efficiency, particularly due to specialized volumetric rendering algorithms mismatched to widely deployed graphics hardware. These computationally intensive methods often necessitate extended rendering times and substantial resources, hindering real-time applications. To address this, exploring alternative data structures or rendering techniques, especially those suitable for low computational mobile devices, presents a promising avenue. \eg  3D Gaussian Splatting~\cite{kerbl20233d} (3DGS) with unprecendeted rendering efficiency and other non-semantically aware strategies could similarly serve as backbone models to SRFs.    
    
    \item [vi] \textbf{Ethical Concerns and Societal Impact.} 
    The generative capabilities of editable NeRFs, allowing for the creation of photorealistic 3D objects, humans, and scenes not previously seen, may lead to challenges akin to those seen with DeepFakes~\cite{shiohara2022detecting} in 2D image generation. These potential issues require similar scrutiny and efforts to mitigate. Conversely, the generative and editable nature of these methods could offer substantial opportunities for 3D enthusiasts and content creators, attributed to their user-friendly design.
    
    \item [vii] \textbf{Performance Evaluation.} Current metrics, such as those used for novel view synthesis, are well-established in the field. However, these metrics are decoupled from human perception, meaning that quantitative evaluations cannot guarantee objective optimality in a way that aligns with human assessment. Multi-task models also face issues due to the absence of composite metrics, and instead rely on disjoint, linear combination metrics. Existing learned perceptual metrics, \eg LPIPS, are restricted to the evaluation of static image frames~\cite{zhang2018unreasonable}. They do not consider video or 3D consistency~\cite{liang2023perceptual} in terms of shape, appearance, and semantics. This is a necessary research opportunity to evaluate complex 3D environments that are typically dynamic.
    
    \item [viii] \textbf{Collaborative Frameworks.} Recognizing the difficulties arising from scattered codebases and the lack of consolidated support, Nerfstudio~\cite{tancik2023nerfstudio} is an end-to-end framework which aggregates modular plug-and-play components, \eg viewers, algorithms, datasets, and benchmarking tools. This facilitates the integration of features across diverse implementations, simplifies the collaborative process for researchers and practitioners, and enhances accessibility through real-time visualization tools that natively support semantic information. Providing a cohesive and extensible platform, such frameworks can foster collaboration, accelerate progress, and contribute to the advancement of NeRF research more efficiently and cohesively.
\end{itemize}

\section{Conclusion}
\label{sec:conclusion}
We have conducted the first survey on Neural Radiance Fields (NeRFs), specifically focusing on semantically-aware NeRFs. Our comprehensive review has shed light on the \sota methodologies, challenges, and a wide array of applications. It also highlights the need for further advancements in this field, to enable more sophisticated, efficient, and context-aware interpretations of 3D scenes by achieving the full potential of NeRFs. This will pave the way for truly real-time end-to-end applications from acquisition to rendering, on commodity hardware.

\nocite{*}

\renewcommand*{\bibfont}{\normalfont\small}
\printbibliography

\end{document}